\documentclass[journal]{IEEEtran}

\usepackage{cite}
\usepackage{amssymb}
\usepackage{amsmath}
\usepackage{epsfig}
\usepackage{graphicx}
\usepackage{amssymb}
\usepackage{amsfonts}
\usepackage{colortbl}
\usepackage{multirow}
\usepackage{longtable}
\usepackage{bm}
\usepackage{multicol}
\usepackage{algorithmic}
\usepackage{amsmath}
\usepackage{amsthm}
\usepackage[linesnumbered,ruled,vlined]{algorithm2e}
\usepackage{mathtools}
\usepackage[dvipsnames,table]{xcolor}
\usepackage{subfigure}
\usepackage{paralist}
\usepackage{color}
\usepackage[hidelinks]{hyperref}

\usepackage{tabu}
\usepackage{rotating}
\usepackage{pifont}
\usepackage{threeparttable}
\usepackage{moresize}
\usepackage{cancel}
\usepackage{pgf-pie}
\usepackage{adjustbox}
\usepackage{url}
\usepackage{subcaption}
\usepackage{booktabs}
\usepackage{array}
\usepackage{makecell}
\usepackage{float}
\usepackage{setspace}
\usepackage{enumitem}
\usepackage{comment}
\usepackage{epstopdf}

\usepackage{booktabs,multirow}
\usepackage{pdflscape}

\usepackage{xcolor}

\newcolumntype{L}{>{\raggedleft}p{0.14\textwidth}}
\newcolumntype{R}{p{0.8\textwidth}}

\newtheorem*{example*}{Example}

\usepackage{xr}
\title{PORTAL: Controllable Landscape Generator for Continuous Optimization -- Part I: Framework}
\author{Danial~Yazdani$^{1,*}$, Mai~Peng$^{2,*}$, Delaram~Yazdani$^{3,*}$, Shima F. Yazdi$^{4}$, Mohammad Nabi Omidvar$^{5}$, Yuan Sun$^{6}$, Trung Thanh Nguyen$^{3,\dagger}$, Changhe Li$^{7,\dagger}$, and Xiaodong Li$^{1,\dagger}$
\thanks{This work was supported in part by the National Natural Science Foundation of China under Grant 62476006, the Hubei Provincial Natural Science Foundation of China under Grant 2023AFA049, the Fundamental Research Funds of the AUST under Grant 2024JBZD0007, and a Liverpool John Moores University Vice-Chancellor PhD Scholarship.}
\thanks{$^{1}$Danial Yazdani and Xiaodong Li are with the School of Computing Technologies, RMIT University, Melbourne, Australia (e-mails: danial.yazdani@gmail.com, xiaodong.li@rmit.edu.au).}%
\thanks{$^{2}$Mai Peng is with the School of Automation, China University of Geosciences, Wuhan, China, and also with the Hubei Key Laboratory of Advanced Control and Intelligent Automation for Complex Systems, Wuhan, China, and also with the Engineering Research Center of Intelligent Technology for Geo-Exploration, Ministry of Education, Wuhan, China. (e-mail: pengmai1998@gmail.com).}%
\thanks{$^{3}$Delaram Yazdani and Trung Thanh Nguyen are with the Liverpool Logistics, Offshore and Marine (LOOM) Research Institute, Faculty of Health, Innovation, Technology and Science, Liverpool John Moores University, Liverpool, United Kingdom (e-mails: delaram.yazdani@yahoo.com, T.T.Nguyen@ljmu.ac.uk).}
\thanks{$^{4}$Shima F. Yazdi is with the RATA Trading Co, Mashhad, Iran (email: mahya.farshchian@gmail.com).}
\thanks{$^{5}$Mohammad Nabi Omidvar is with the School of Computing, University of Leeds, and Leeds University Business School, Leeds, United Kingdom. (e-mail: m.n.omidvar@leeds.ac.uk)}
\thanks{$^{6}$Yuan Sun is with the Research Center for Data Analytics and Cognition, La Trobe University, Melbourne, Australia. (email: Yuan.Sun@latrobe.edu.au)}
\thanks{$^{7}$Changhe Li is with the School of Artificial Intelligence, Anhui University of Science \& Technology, State Key Laboratory of Digital Intelligent Technology for Unmanned Coal Mining, Anhui University of Science \& Technology, Hefei, China. (email: changhe.lw@gmail.com)}
\thanks{$^{*}$Equal Contribution}
\thanks{$^{\dagger}$Corresponding authors}
}

\begin{document}
%\markboth{IEEE Transactions on Evolutionary Computation}%
%{Yazdani~\MakeLowercase{\textit{et al.}}: Robust Optimization Over Time: In-depth Technical Review, Taxonomy, and Future Directions}
\maketitle
%\linenumbers

\begin{abstract}
Benchmarking is central to optimization research, yet existing test suites for continuous optimization remain limited: classical collections are fixed and rigid, while previous generators cover only narrow families of landscapes with restricted variability and control over details. 
This paper introduces \textbf{PORTAL} (\textit{P}latform for \textit{O}ptimization \textit{R}esearch, \textit{T}esting, \textit{A}nalysis, and \textit{L}earning), a general benchmark generator that provides fine-grained, independent control over basin curvature, conditioning, variable interactions, and surface ruggedness. 
PORTAL's layered design spans from individual components to block-wise compositions of multi-component landscapes with controllable partial separability and imbalanced block contributions.
It offers precise control over the shape of each component in every dimension and direction, and supports diverse transformation patterns through both element-wise and coupling operators with compositional sequencing. 
All transformations preserve component centers and local quadratic structure, ensuring stability and interpretability. 
A principled neutralization mechanism prevents unintended component domination caused by exponent or scale disparities, which addresses a key limitation of prior landscape generators. 
On this foundation, transformations introduce complex landscape characteristics, such as multimodality, asymmetry, and heterogeneous ruggedness, in a controlled and systematic way. 
PORTAL enables systematic algorithm analysis by supporting both isolation of specific challenges and progressive difficulty scaling. 
It also facilitates the creation of diverse datasets for meta-algorithmic research, tailored benchmark suite design, and interactive educational use.
The complete Python and MATLAB source code for PORTAL is publicly available at  
\href{https://github.com/EvoMindLab/PORTAL}{\texttt{[https://github.com/EvoMindLab/PORTAL]}}.

\end{abstract}

\IEEEpeerreviewmaketitle

 \section{Introduction}
 \label{sec:Introduction}

\IEEEPARstart{O}{ptimization} pervades virtually every domain of science and engineering, from machine learning and data analysis to aerospace design and financial modeling~\cite{bottou2018optimization,li2023evaluation,cao2022ai}. 
As the complexity and scale of these applications grow, so does the need for robust, well-understood algorithms that can navigate increasingly challenging problem landscapes. 
Meaningful progress depends on our ability to systematically evaluate, compare, and \emph{understand} algorithmic behavior under controlled conditions, a level of insight that theoretical analysis alone may not provide given the complexity of modern algorithms and problem classes.

Synthetic benchmarking provides this empirical foundation~\cite{bartz2020benchmarking}. 
Because benchmark objectives are explicitly defined, they offer known structural characteristics and known optima (location and value). 
This transparency enables reproducible experiments, unambiguous success criteria, and fair cross-study comparisons. 
Well-designed benchmarks also exhibit essential qualities: \emph{diversity} across problem types, \emph{complexity variety} spanning unimodal to highly multimodal and deceptive cases, \emph{algorithmic neutrality} that avoids bias toward specific methods, and \emph{practicality} through clear interpretation of algorithm behavior~\cite{whitley2002testing,shir2018compiling,li2018open}. 
They are naturally scalable in dimension and problem size, and their accessibility through open code and documentation ensures broad adoption. 
Where benchmark \emph{families} expose parameters, one can systematically vary difficulty and perform controlled ablations; where they do not, the availability of ground truth still permits clean, repeatable evaluation. 

Beyond single-algorithm studies, synthetic benchmarks also underpin the emerging field of \emph{meta-algorithmics}: datasets of labeled instances, either by design or via feature extraction (e.g., Exploratory Landscape Analysis~\cite{munoz2014exploratory}), support automated algorithm selection~\cite{tornede2023algorithm,prager2022automated} and algorithm design~\cite{zhao2023automated,liu2024systematic}.

Despite their long-standing utility, existing benchmarking resources remain fundamentally constrained. 
Classical collections, CEC competition suites~\cite{suganthan2005problem,yue2019problem}, and the BBOB framework~\cite{hansen2009real} are all \emph{fixed and finite}: they provide valuable reference problems but offer little parameterization, making it difficult to isolate individual characteristics such as conditioning or multimodality. 
Their coverage of interaction structures and ruggedness patterns is limited, and they cannot systematically generate new instances beyond the predefined set.  

Benchmark generators such as the Generalized Numerical Benchmark Generator (GNBG)~\cite{gandomi2023gnbg}, which is a static extension of the dynamic Generalized Moving Peaks Benchmark (GMPB)~\cite{yazdani2020benchmarking,yazdani2021ieee}, introduced broader parametric variation than fixed suites, but important gaps remain.
Key features are often entangled; for example, separability is tied to exponent choices, preventing clean control of variable interactions. 
Ruggedness and multimodality arise only from narrow transformation families, restricting the diversity of local-optima patterns that can be produced. 
Random instance generation is further undermined by bias, as exponent and scale effects cause low-exponent components to dominate landscapes. 
Moreover, existing generators provide little systematic support for constructing partially separable structures or composed challenges.  

These shortcomings are particularly limiting for emerging \emph{meta-algorithmic} research, where unbiased random generation and broad pattern diversity are essential for building large, representative training datasets. 
Current generators cannot deliver the coverage and variety required, reinforcing the need for a more advanced and flexible framework.

To address these limitations, we introduce \emph{PORTAL}: a \emph{P}latform for \emph{O}ptimization \emph{R}esearch, \emph{T}esting, \emph{A}nalysis, and \emph{L}earning.
PORTAL is a general benchmark generator designed to enable highly flexible construction of continuous optimization problems. 
Its architecture resolves the entanglements and biases of prior generators while greatly expanding the range of characteristics and diversity of landscapes that can be produced.

PORTAL is organized into three conceptual layers:
\begin{itemize}
    \item \emph{Component layer}: Defines the structure of individual basins with independent control over curvature, anisotropy, conditioning, and variable interactions. 
    At this level, users can configure both \emph{dimension-wise} and \emph{directional} characteristics through independent scaling factors and exponents, offering a degree of flexibility absent in prior generators. 
    Component-wise transformations, whether element-wise or coupling and applied individually or in sequence, further inject multimodality, ruggedness, and asymmetry while preserving the component center and local quadratic structure.
    Four of the available transformations support this dimension-wise and directional control, enabling asymmetric configurations and diverse patterns not achievable in previous frameworks, while the remaining radial transformations contribute complementary isotropic effects.
    \item \emph{Landscape layer}: Constructs landscapes by combining multiple components via the $\min$ operator. This layer enables systematic control over the number, location, and relative quality of basins, supporting multimodal and deceptive problem settings.  
    \item \emph{Composition layer}: Assembles full benchmark problems by combining multiple landscapes, enabling partial separability and weighted contributions (including intentional contribution imbalance).
    This enables the creation of complex composite structures while maintaining interpretability.  
\end{itemize}
Together, these layers provide fine-grained, independent control over all major landscape characteristics. 
By resolving entangled parameters, supporting systematic composition, and enabling diverse, unbiased random instance generation, PORTAL provides the foundation not only for fair and rigorous algorithm evaluation, but also for emerging applications such as meta-algorithmics, competition suite design, and educational demonstrations.

The remainder of this paper is organized as follows. 
Section~\ref{sec:relatedWork} reviews related work and motivates the need for a new benchmark generator. 
Section~\ref{sec:portal} introduces the PORTAL framework in detail, describing its layered architecture, component structures, and transformation operators. 
Section~\ref{sec:applications} illustrates key applications and use cases of PORTAL, ranging from systematic algorithm analysis to benchmark suite generation, meta-algorithmic datasets, and educational demonstrations. 
Finally, Section~\ref{sec:conclusion} concludes the paper.

\section{Related Work and Motivation}
\label{sec:relatedWork}

Benchmark test functions are indispensable for objectively evaluating and comparing optimization algorithms. In the domain of unconstrained, single objective continuous optimization, which is the focus of this paper, three major resources have become dominant in the literature: collections of classical functions~\cite{jamil2013literature,hussain2017common}, the IEEE CEC competition suites~\cite{suganthan2005problem,yue2019problem}, and the BBOB (Black Box Optimization Benchmarking) suite~\cite{hansen2009real}.
More recently, parametric benchmark generators~\cite{li2018open,gandomi2023gnbg} have emerged as a fourth category, offering new capabilities for systematic benchmark construction. Below, we briefly review these benchmark categories as the most commonly used and most relevant to the scope of this work.

\subsection{Commonly Used Benchmark Functions}

\subsubsection{Classical Mathematical Functions}

Classical benchmark functions provide the mathematical foundation underlying modern benchmark design. 
This collection includes well-established functions such as Sphere, Rosenbrock, Rastrigin, Ackley, and Griewank, each exhibiting distinct landscape characteristics: the Sphere function tests convergence with its convex unimodal structure; Rosenbrock challenges algorithms with narrow parabolic valleys; and Rastrigin evaluates global search capability through its highly multimodal landscape.
A comprehensive survey by Jamil and Yang~\cite{jamil2013literature} cataloged 175 distinct functions, systematically classified by modality, separability, dimensional scalability, and landscape ruggedness. 
These functions remain essential for algorithm validation, theoretical analysis, and educational use.
Moreover, they form the mathematical building blocks for modern CEC and BBOB functions, which extend classical landscapes through composition, rotation, and sophisticated transformations.

\subsubsection{IEEE CEC Competition Suites}

Since 2005, the IEEE Congress on Evolutionary Computation (CEC) has released benchmark suites that serve as official testbeds for algorithmic competitions, becoming the de facto standard for evaluating evolutionary and metaheuristic optimization algorithms\footnote{The CEC community has released benchmark suites for several problem domains, including large-scale optimization~\cite{tang2007benchmark,li2013benchmark} and multimodal optimization (with multiple global optima)~\cite{li2013multimodalbenchmark}, among others. 
In this paper, we focus exclusively on unconstrained, single-objective continuous global optimization.}
These suites are primarily built upon well-established classical mathematical functions, as well as their transformed variants and compositions. 
Transformations commonly include shifting, rotation, and biasing, which aim to increase problem complexity and reduce structural bias.
Many benchmark problems also involve hybridization and composition of multiple base functions, both transformed and untransformed, to construct more challenging landscapes.

CEC 2005~\cite{suganthan2005problem} introduced 25 benchmark functions, establishing the foundation with categories including unimodal, multimodal, hybrid, and composition functions.  
CEC 2014--2017~\cite{chen2014problem,awad2016problem} refined the suite to 30 functions, incorporating more sophisticated hybrid and composition designs, along with enhanced transformations to increase landscape complexity.  
CEC 2020--2024~\cite{mohamed2020problem} retained the 30-function format while introducing additional transformations.

\subsubsection{BBOB (Black-Box Optimization Benchmarking) Suite}

The BBOB suite, developed by Hansen et al.~\cite{hansen2009real}, represents one of the most methodologically rigorous benchmarking frameworks for continuous optimization. 
Integrated within the COCO (Comparing Continuous Optimizers) platform~\cite{hansen2021coco}, BBOB comprises 24 scalable functions systematically organized into five problem categories: separable functions, moderately conditioned functions, ill-conditioned functions, multimodal functions with adequate global structure, and multimodal functions with weak global structure.

Each function supports various dimensionalities and includes a number of randomized instances per function, enabling statistically robust performance evaluation across subtly different but isomorphic problem variants. 
All functions employ sophisticated transformations, including rotation matrices, translation vectors, and asymmetric distortions, in order to eliminate coordinate system dependencies and algorithm-specific biases.

\subsubsection{Benchmark Generators}

Benchmark generators have been particularly influential in the field of optimization in dynamic environments~\cite{yazdani2021DOPsurveyPartA,yazdani2021DOPsurveyPartB}, where algorithms must adapt to landscapes that change over time. 
Dynamic optimization problems are characterized by objective functions, constraints, or optima that vary during the optimization process, requiring algorithms to track moving optima and adapt to environmental changes. 
Because systematic generation and controllable modifications of the landscape are essential in this context, parametric generators emerged as a natural solution.

The Moving Peaks Benchmark (MPB)~\cite{branke1999memory} was one of the earliest and most widely adopted generators. 
In its original form, MPB modeled the landscape as a set of $m$ peaks whose positions, heights, and widths vary over time:
\begin{align}
f^{(t)}(\mathbf{x}) = \max_{i \in \{1,\ldots,m\}} \frac{h_i^{(t)}}{1 + w_i^{(t)} \sum_{j=1}^{d}(x_j - c_{i,j}^{(t)})^2},
\end{align}
where $h_i^{(t)}$, $w_i^{(t)}$, and $\mathbf{c}_i^{(t)}$ denote the height, width, and center of the $i$-th peak at time $t$. 
Later variants replaced the quadratic basin with cone-shaped basins~\cite{branke2003designing}:
\begin{align}
f^{(t)}(\mathbf{x}) = \max_{i \in \{1,\ldots,m\}} \left\{ h_i^{(t)} - w_i^{(t)} \left\|\mathbf{x} - \mathbf{c}_i^{(t)}\right\| \right\},
\end{align}
which became the de facto standard configuration~\cite{yazdani2021DOPsurveyPartB}.

The Generalized Moving Peaks Benchmark (GMPB)~\cite{yazdani2020benchmarking} extended MPB in three important directions. 
First, it introduced diagonal weight matrices that allow anisotropic scaling, producing elliptical rather than spherical basins. 
Second, it added non-linear transformations to inject local irregularities and ruggedness within each basin. 
Third, it employed Givens rotation matrices to control variable interactions and enable additional dynamic changes through time-varying rotations. 
The resulting formulation is:
\begin{align}
\resizebox{0.98\columnwidth}{!}{$
f^{(t)}(\mathbf{x}) = \max_{i \in \{1,\ldots,m\}} \left\{ h_i^{(t)} - \sqrt{\left(\mathbf{x} - \mathbf{c}_i^{(t)}\right)^T \mathbf{R}_i^{(t)} \mathbf{W}_i^{(t)} \mathbf{T}\!\left(\mathbf{R}_i^{(t)}\left(\mathbf{x} - \mathbf{c}_i^{(t)}\right)\right)} \right\},$}
\end{align}
where $\mathbf{R}_i^{(t)}$ is a rotation matrix generated by Givens-based rotation matrices, $\mathbf{W}_i^{(t)}$ is a diagonal scaling matrix, and $\mathbf{T}(\cdot,i)$ is an element-wise non-linear transformation defined as
\begin{align}
\resizebox{0.98\columnwidth}{!}{$
\mathbf{T}(y_j, i) = \begin{cases}
\exp\!\left(\log(y_j) + \tau_i^{(t)}\!\left[\sin\!\left(\omega_{i,1}^{(t)} \log(y_j)\right) + \sin\!\left(\omega_{i,2}^{(t)} \log(y_j)\right)\right]\right) & \text{if } y_j > 0, \\[4pt]
0 & \text{if } y_j = 0, \\[4pt]
-\exp\!\left(\log(|y_j|) + \tau_i^{(t)}\!\left[\sin\!\left(\omega_{i,3}^{(t)} \log(|y_j|)\right) + \sin\!\left(\omega_{i,4}^{(t)} \log(|y_j|)\right)\right]\right) & \text{if } y_j < 0,
\end{cases}$}
\end{align}
with $\tau_i^{(t)}$ controlling amplitude and $\omega_{i,j}^{(t)}$ tuning ruggedness frequency.

While powerful, GMPB presents two notable limitations. 
First, the enforced square root (power exponent of $\frac{1}{2}$) makes all components inherently non-separable, reducing rotations to a source of temporal variation rather than genuine structural diversity. 
Second, the log-sinusoidal transformation produces a fixed, repetitive irregularity pattern with limited capacity to simulate diverse local optima topologies.

Building on this foundation, the Generalized Numerical Benchmark Generator (GNBG)~\cite{gandomi2023gnbg} can be regarded as a static specialization of GMPB. 
It removed temporal dynamics, switched the task to minimization, and replaced the fixed square root with a configurable exponent parameter $\lambda_k$ for each component, enabling direct control over basin curvature. 
Although GNBG marks a step toward parametric benchmark generation in static continuous optimization, it still inherits GMPB's constraints on separability and irregularity diversity.

A recent generator, MA-BBOB~\cite{vermetten2025mabbob}, extends the BBOB suite by constructing affine combinations of existing BBOB functions with random shifts and logarithmic value rescaling. 
This design allows smooth interpolation between BBOB instances and generation of large sets of related instances for algorithm-selection and AutoML
studies. 
Unlike GNBG, MA-BBOB does not define landscapes from first principles or expose explicit control over geometric properties such as curvature, conditioning, or ruggedness. 
Instead, it reuses the BBOB primitives as building blocks, offering diversity through mixture weights rather than parametric structure.

\subsection{Limitations of Existing Approaches and Motivation for PORTAL}
\label{sec:motivation}

While the benchmark resources described above have enabled significant progress in algorithm evaluation, they present fundamental limitations that hinder comprehensive algorithmic analysis and emerging research directions.

\subsubsection{Fixed Benchmark Limitations} Traditional benchmark suites suffer from inherent rigidity: their characteristics are predefined and immutable, severely limiting researchers' ability to conduct targeted, fine-grained analyses. This inflexibility prevents systematic exploration of fundamental questions such as: At what degree of ill-conditioning does an algorithm's performance collapse? How many local optima can an algorithm navigate before its global search capability breaks down? What variable interaction complexity overwhelms different algorithmic approaches? Without parametric control over these characteristics, researchers cannot isolate specific challenges or understand algorithmic trade-offs across the landscape feature spectrum.

\subsubsection{Generator Limitations}
While GNBG represented a step toward parametric benchmark generation, more recent works such as MA-BBOB introduced broader statistical combinations of existing functions. 
Despite their different formulations, both frameworks face limitations that constrain their utility for comprehensive evaluation and modern meta-algorithmic applications. 
MA-BBOB extends the BBOB suite and offers statistical diversity rather than structural control. 
While valuable for producing large sets of related instances, it does not provide explicit parametric access to fundamental geometric properties such as curvature, conditioning, separability, or ruggedness, and therefore cannot serve as a general-purpose benchmark constructor. 
GNBG, on the other hand, was intended as a first-principle, general-purpose generator but exhibits several key shortcomings:

\begin{itemize}
\item \emph{Interaction Control Constraints:} GNBG inherited GMPB's fundamental issue where exponents other than unity render components inherently non-separable, regardless of rotation matrix configuration. This severely limits control over variable interaction structures, a crucial capability for systematic algorithm analysis.

\item \emph{Random Generation Challenges:} GNBG's design induces systematic dominance effects, where components with smaller exponents overwhelm those with larger exponents in multi-component landscapes. This makes principled random instance generation nearly impossible, as landscapes become skewed toward lower-exponent components.

\item \emph{Restricted Topological Diversity:} The reliance on a narrow set of transformations (e.g., log-sinusoidal) produces relatively fixed irregularity patterns and restricts the variety of basin shapes, local optima arrangements, and ruggedness profiles that can be generated. As a result, GNBG cannot provide the breadth of landscape diversity required for comprehensive algorithm evaluation.

\end{itemize}

\subsubsection{The Meta-Algorithmic Imperative} Perhaps most critically, the emergence of meta-algorithmics, including algorithm
selection~\cite{tornede2023algorithm} and automated algorithm
design~\cite{zhao2023automated,liu2024systematic}, demands a paradigm shift in
benchmark generation.
These approaches require access to vast, diverse datasets of problem instances with controlled yet varied characteristics. 
Traditional fixed suites and existing generators fall short of this requirement, lacking both the volume and systematic diversity needed to train robust, generalizable meta-algorithmic systems.

\subsubsection{The PORTAL Solution}
In response, this paper introduces PORTAL, a comprehensive framework designed to address a broad range of benchmark generation challenges. PORTAL provides:

\paragraph{High-Resolution Control} Fine-grained and largely independent control over key landscape characteristics, including variable interactions, conditioning, local optima patterns, and asymmetries.

\paragraph{Principled Random Generation} Mechanisms for producing diverse and unbiased random instances suitable for meta-algorithmic training and evaluation.

\paragraph{Extensive Instance Diversity} A parametric design that supports the creation of a wide variety of unique instances spanning a broad spectrum of optimization challenges.

\paragraph{Layered Architecture} A compositional framework allowing systematic construction of simple to extremely complex optimization scenarios.

\paragraph{Meta-Learning Ready} Capabilities tailored for generating large-scale, diverse training and evaluation datasets to support next-generation meta-algorithmic research.

PORTAL enables researchers to systematically explore the space of optimization landscapes and deepen our understanding of algorithmic behavior across a wide range of continuous optimization settings.

\section{PORTAL Framework}
\label{sec:portal}

\subsection{Baseline}
\label{sec:baseline}

PORTAL's baseline is organized hierarchically: Component functions define the local shape of components (Layer 1). A landscape is then formed as the lower envelope of multiple such components (Layer 2). Finally, composite landscapes are built by combining several landscapes on disjoint subsets of variables (Layer 3).

\subsubsection{Layer 1: Component Functions}
\label{sec:layer1}

\paragraph{Per-Direction Exponent Formulation (First Component Function)}
\label{sec:Form1}

Consider a component function indexed by $k$, centered at $\mathbf{c}_k \in \mathbb{R}^d$, with a rotation matrix $\mathbf{R}_k \in \mathbb{R}^{d \times d}$ and an optional nonlinear transformation $\mathfrak{T}_k:\mathbb{R}^d \to \mathbb{R}^d$. For an input $\mathbf{x}\in\mathbb{R}^d$, the internal coordinates of component $k$ are
\begin{align}
\label{eq:InternalCoordinate}
\mathbf{z}_k &= \mathfrak{T}_k\!\bigl(\mathbf{R}_k(\mathbf{x}-\mathbf{c}_k)\bigr).
\end{align}
The mapping $\mathfrak{T}_k$ is a nonlinear transformation applied to the internal coordinates to create local optima and ruggedness within each component's basin. 
When $\mathfrak{T}_k$ is the identity transformation, the component has a smooth basin. 
Rotation matrix generation is detailed in Section~\ref{sec:rotation}, and available transformations are described in Section~\ref{sec:transforms}.

Each dimension $i$ has anisotropy factors $\kappa_{k,i}^{+}$ and $\kappa_{k,i}^{-}$ (collectively denoted as $\kappa_{k,i}^{\pm}$) for the positive and negative directions, respectively, with $\kappa_{k,i}^{\pm} > 0$. 
Each direction also has its own exponent $\mathfrak{p}_{k,i}^{\pm}>0$, and $\beta_k \in \mathbb{R}$ is the vertical offset. The first component function is
\begin{align}
\label{eq:Form1}
\phi^{(1)}_k(\mathbf{x}) = \beta_k \;+\; \sum_{i=1}^{d} \rho^{\mathrm{sgn}(z_{k,i})}_{k,i}\,
\kappa_{k,i}^{\,\mathrm{sgn}(z_{k,i})}\,|z_{k,i}|^{\,2\,\mathfrak{p}_{k,i}^{\,\mathrm{sgn}(z_{k,i})}},
\end{align}
where $\mathrm{sgn}(\cdot)\in\{-,+\}$ selects the direction-specific parameters and $\rho^{\pm}_{k,i}>0$ are per-term neutralizing factors.
We use the exponent $2\,\mathfrak{p}_{k,i}^{\pm}$ rather than $\mathfrak{p}_{k,i}^{\pm}$ so that $\mathfrak{p}=1$ recovers the standard quadratic basin $z^2$; values $\mathfrak{p}<1$ and $\mathfrak{p}>1$ then naturally produce subquadratic and superquadratic curvature, respectively.

The per-direction exponents $\mathfrak{p}_{k,i}^{\pm}$ provide fine-grained control over curvature along each coordinate direction: $\mathfrak{p}_{k,i}^{\pm}=0.5$ yields \emph{linear} growth in that direction, $\mathfrak{p}_{k,i}^{\pm}<0.5$ is \emph{sublinear}, and $\mathfrak{p}_{k,i}^{\pm}>0.5$ is \emph{superlinear}. 
This allows different curvature profiles for each dimension-direction pair.
The anisotropy factors $\kappa_{k,i}^{\pm}$ control basin elongation and conditioning. 

If no corrective mechanism is applied, both the exponents and anisotropy factors would affect basin size, creating unintended imbalances in multi-component landscapes. 
Components with smaller exponents grow more slowly away from their centers, producing broader basins that can overshadow components with larger exponents when combined via the $\min$ operator (Layer 2). Similarly, larger anisotropy factors would amplify basin extent along specific directions, conflating shape control with scale control. To prevent these effects and enable independent control of shape versus scale, we introduce a neutralization mechanism.

We define a parameter $\Delta>0$ that directly controls the vertical rise of the component at a canonical reference radius $r_{\mathrm{ref}}$. 
The neutralizing factors $\rho^{\pm}_{k,i}$ are chosen to compensate for both exponent and anisotropy effects, ensuring that the component rises by $\Delta$ at the reference point regardless of the chosen $\mathfrak{p}_{k,i}^{\pm}$ and $\kappa_{k,i}^{\pm}$ values. 
This separates parameter roles: $\mathfrak{p}_{k,i}^{\pm}$ controls curvature, $\kappa_{k,i}^{\pm}$ controls anisotropy and conditioning, and $\Delta$ controls scale.

Let $\bar{\kappa}_{k,i} = \tfrac{1}{2}(\kappa_{k,i}^+ + \kappa_{k,i}^-)$ denote the average anisotropy for dimension $i$, and define the component-level reference as the arithmetic mean
\begin{align}
\bar{\kappa}_k = \frac{1}{d}\sum_{i=1}^{d}\bar{\kappa}_{k,i}.
\end{align}
At the reference radius, ignoring rotation and transformation, each dimension contributes a canonical term $\bar{\kappa}_k \cdot r_{\mathrm{ref}}^{2\mathfrak{p}_{k,i}^{\pm}}$. Requiring the total function value to rise by $\Delta$ from the center, each term should contribute $\Delta/d$:
\begin{align}
\label{eq:Form1Neutralization}
\rho^{\mathrm{sgn}(z_{k,i})}_{k,i}\, \bar{\kappa}_k \cdot r_{\mathrm{ref}}^{2\,\mathfrak{p}_{k,i}^{\mathrm{sgn}(z_{k,i})}} & = \frac{\Delta}{d}
\quad\Longrightarrow\quad \nonumber \\
& \rho^{\mathrm{sgn}(z_{k,i})}_{k,i} = \frac{\Delta/d}{\bar{\kappa}_k \cdot r_{\mathrm{ref}}^{2\,\mathfrak{p}_{k,i}^{\mathrm{sgn}(z_{k,i})}}}\;.
\end{align}
With this neutralization, the exponents and anisotropy factors define only the shape and conditioning of the basin, while $\Delta$ controls its vertical scale. The offset $\beta_k$ sets the function value at the component center, so $\phi^{(1)}_k(\mathbf{c}_k)=\beta_k$. 
In this formulation, the conditioning is determined solely by the ratio between the largest and smallest anisotropy factors $\kappa_{k,i}^{\pm}$, whereas the exponents affect only the curvature profile without influencing axis elongation.
Figure~\ref{fig:form1} illustrates how variations in the per-direction exponents and anisotropy parameters in~\eqref{eq:Form1} shape the geometry of the first component function.

\paragraph{Single-Exponent Formulation (Second Component Function)}
\label{sec:Form2}

This alternative formulation uses a single power exponent $\mathfrak{p}_k>0$ per component, applied to the sum of weighted quadratic terms rather than to individual terms. The internal coordinates $\mathbf{z}_k$ are computed as in the first component function using Equation~\eqref{eq:InternalCoordinate}. With anisotropy factors $\kappa_{k,i}^{\pm}>0$ and vertical offset $\beta_k \in \mathbb{R}$, the second component function is defined as
\begin{align}
\label{eq:Form2}
\phi^{(2)}_k(\mathbf{x}) = \beta_k \;+\; \rho_k \Bigg(\sum_{i=1}^{d} 
   \bigl(\kappa_{k,i}^{\,\mathrm{sgn}(z_{k,i})}\bigr)^{1/\mathfrak{p}_k}\,|z_{k,i}|^{2}
\Bigg)^{\mathfrak{p}_k},
\end{align}
where $\mathrm{sgn}(\cdot)\in\{-,+\}$ selects the direction-specific anisotropy and $\rho_k>0$ is a neutralizing factor.
The exponent $\mathfrak{p}_k$ governs the curvature of the entire basin: $\mathfrak{p}_k=0.5$ yields \emph{linear} growth in distance, $\mathfrak{p}_k<0.5$ is \emph{sublinear}, and $\mathfrak{p}_k>0.5$ is \emph{superlinear}. 
The anisotropy factors $\kappa_{k,i}^{\pm}$ control basin elongation and conditioning.

Following the neutralization approach established for the first component function, we compensate for both exponent and anisotropy effects. Define averaged anisotropy factors
\begin{align}
\bar{\kappa}_{k,i} = \tfrac{1}{2}(\kappa_{k,i}^+ + \kappa_{k,i}^-),
\end{align}
and compute the canonical quadratic mass at radius $r_{\mathrm{ref}}$ as
\begin{align}
s_k^{\star} = r_{\mathrm{ref}}^{2}\sum_{i=1}^{d} \bar{\kappa}_{k,i}^{1/\mathfrak{p}_k}.
\end{align}
Requiring the powered term to rise by $\Delta>0$ at this reference point yields
\begin{align}
\label{eq:Form2Neutralization}
\rho_k\, \bigl(s_k^{\star}\bigr)^{\mathfrak{p}_k} = \Delta
\quad\Longrightarrow\quad
\rho_k = \frac{\Delta}{\Bigl(r_{\mathrm{ref}}^{2}\sum_{i=1}^{d} \bar{\kappa}_{k,i}^{1/\mathfrak{p}_k}\Bigr)^{\mathfrak{p}_k}}\;.
\end{align}
This neutralization achieves the same separation of concerns as in the first component function: $\mathfrak{p}_k$ controls curvature, $\kappa_{k,i}^{\pm}$ controls shape and conditioning, and $\Delta$ controls scale. 
The explicit factor $(\kappa_{k,i}^{\pm})^{1/\mathfrak{p}_k}$ inside the sum ensures that the exponent $\mathfrak{p}_k$ does not distort relative axis scales after raising to the power and applying neutralization. As a result, the conditioning is determined solely by the ratio between the largest and smallest anisotropy factors $\kappa_{k,i}^{\pm}$, whereas the exponent $\mathfrak{p}_k$ affects only curvature without influencing axis elongation. The offset $\beta_k$ sets the function value at the component center, so $\phi^{(2)}_k(\mathbf{c}_k)=\beta_k$.
A corresponding visualization of the single-exponent formulation for varying  $\kappa_{k,i}^{\pm}$ values is provided in Figure~\ref{fig:form2}.

\subsubsection{Layer 2: Component-assembled Landscape}
\label{sec:layer2}

A landscape is constructed by combining multiple component functions through a selection operator. 
Given a set of components $\mathcal{C}$ where $|\mathcal{C}|$ denotes the number of components, the landscape function is defined as\footnote{Note that throughout this paper, all formulations and discussions are presented in the \emph{minimization} setting for consistency. For maximization instances, the $\min$ operator in Equation~\eqref{eq:Layer2Baseline} is replaced with $\max$, and the component functions are modified by changing the sign of the powered contribution (i.e., replacing ``$+$" with ``$-$" preceding the $\rho$ terms in Equations~\eqref{eq:Form1} and~\eqref{eq:Form2}). This transforms valleys into peaks while preserving all curvature and scaling properties.}
\begin{align} 
\label{eq:Layer2Baseline} 
f(\mathbf{x}) = \min_{k \in \mathcal{C}} \phi^{(n_k)}_k(\mathbf{x}), 
\end{align} 
where $n_k \in \{1,2\}$ indicates whether component $k$ uses the first or second component function formulation from Sections~\ref{sec:Form1} and~\ref{sec:Form2}, respectively. 
This allows mixing different component types within a single landscape.

The $\min$ operator partitions the landscape: each point $\mathbf{x}$ belongs to the basin of attraction of the component that achieves the minimum value at that location. 
The resulting landscape exhibits multiple distinct basins of attraction, with the global minimum occurring at the center of the component with the lowest offset value $\beta_k$.
Setting $|\mathcal{C}|=1$ reduces to a single component landscape.
A visual illustration of a two-component landscape constructed from the first and second component formulations, each using different exponent values, is provided in Figure~\ref{fig:layer2}.

\subsubsection{Layer 3: Block-Wise Landscape Composition}
\label{sec:layer3}

The previous layer defines landscapes based on single or multiple component functions over the full decision space. 
While a single-component landscape can exhibit partial separability through appropriate rotation matrix configuration, any multi-component landscape is inherently fully non-separable. 
To enable modeling of \emph{partially separable} instances with multiple components and heterogeneous substructures, we introduce a block-wise composition mechanism.

Let $b$ denote the number of blocks. 
The decision vector $\mathbf{x} \in \mathbb{R}^d$ is partitioned into index sets $\mathcal{I}_1, \ldots, \mathcal{I}_b$ such that $\bigcup_{i=1}^{b} \mathcal{I}_i = \{1,2,\ldots,d\}$ and $\mathcal{I}_i \cap \mathcal{I}_j = \varnothing$ for $i\neq j$ (blocks are disjoint by default).

Each block $\mathcal{I}_i$ defines a subvector $\mathbf{x}_{\mathcal{I}_i} \in \mathbb{R}^{|\mathcal{I}_i|}$ and corresponds to an independent subfunction $f_i : \mathbb{R}^{|\mathcal{I}_i|} \to \mathbb{R}$, constructed according to Equation~\eqref{eq:Layer2Baseline}. 
Each subfunction may contain one or multiple components with its own parameter settings. 
A positive weight $w_i$ controls the contribution of subfunction $i$ to the composite function:
\begin{align}
F(\mathbf{x}) = \sum_{i=1}^{b} w_i \, f_i\!\bigl(\mathbf{x}_{\mathcal{I}_i}\bigr).
\label{eq:layer3}
\end{align}
Setting $b = 1$ and $w_1 = 1$ recovers the Layer 2 formulation.

Note that PORTAL can also allow blocks to overlap (i.e., $\mathcal{I}_i \cap \mathcal{I}_j \neq \varnothing$ for some $i\neq j$). 
In that case, the sum of per-block minima, $\sum_{i=1}^b w_i \min_{\mathbf{u}\in\mathbb{R}^{|\mathcal{I}_i|}} f_i(\mathbf{u})$,
is generally a \emph{lower bound} rather than an attainable value unless all block minimizers agree on shared coordinates.

\subsection{Rotation Matrix Generation}
\label{sec:rotation}

PORTAL employs Givens rotation matrices to enable precise control over variable interactions within component basins. While detailed descriptions appear in \cite{cheng2018solving,gandomi2023gnbg}, we provide a treatment here for completeness and self-containment.

The effectiveness of rotation-based interaction control depends on the component's inherent separability properties. 
The first component function (Section~\ref{sec:Form1}) maintains full separability by design and only becomes rotation-invariant when all exponents equal unity, making it suitable for designing controlled variable interactions. 
Conversely, the second component function (Section~\ref{sec:Form2}) becomes rotation-invariant when using identity transformations with uniform $\kappa$ values, and becomes inherently non-separable when $\mathfrak{p} \neq 1$. 
In such cases, rotation matrices can still alter variable couplings but cannot precisely define or control the resulting interaction structure.

The transformations discussed in Section~\ref{sec:transforms} further influence this control: element-wise transformations preserve the existing interaction structure, while coupling transformations introduce full non-separability regardless of rotation settings.

A fundamental Givens rotation in two dimensions is defined as:
\begin{align}
\label{eq:2DimGivens}
\mathbf{G} = \begin{pmatrix}
\cos(\psi) & -\sin(\psi) \\
\sin(\psi) & \cos(\psi)
\end{pmatrix},
\end{align}
where $\psi$ represents the rotation angle. For higher-dimensional spaces, Givens rotations selectively operate within two-dimensional subspaces while preserving all orthogonal dimensions. To establish interaction between variables $u$ and $v$, the $d$-dimensional Givens matrix is constructed as:
\begin{align}
\label{eq:givens}
\bigl[G^{(u,v)}(\psi)\bigr]_{ij} =
\begin{cases}
1, & i=j \text{ and } i\notin\{u,v\},\\[2pt]
\cos\psi, & (i,j)\in\{(u,u),(v,v)\},\\[2pt]
-\sin\psi, & (i,j)=(u,v),\\[2pt]
\ \ \sin\psi, & (i,j)=(v,u),\\[2pt]
0, & \text{otherwise.}
\end{cases}
\end{align}

Consider a practical example involving variable interaction between the second and fifth coordinates in a 6-dimensional space:
\begin{align}
\label{eq:GivensExample}
\mathbf{G} = \begin{pmatrix}
1 & 0 & 0 & 0 & 0 & 0 \\
0 & \cos(\psi_{2,5}) & 0 & 0 & -\sin(\psi_{2,5}) & 0 \\
0 & 0 & 1 & 0 & 0 & 0 \\
0 & 0 & 0 & 1 & 0 & 0 \\
0 & \sin(\psi_{2,5}) & 0 & 0 & \cos(\psi_{2,5}) & 0 \\
0 & 0 & 0 & 0 & 0 & 1
\end{pmatrix}.
\end{align}

Setting $\psi_{u,v} = 0$ yields the identity matrix, preserving the original interaction structure. 
The interaction strength correlates with the angle's deviation from axis-aligned orientations (multiples of $\tfrac{\pi}{2}$). 
For example, $\psi = \tfrac{\pi}{3}$ creates stronger coupling than $\psi = \tfrac{\pi}{12}$.

To construct the complete rotation matrix $\mathbf{R}_k$ for component $k$, an upper-triangular interaction matrix $\boldsymbol{\Psi}_k$ of size $d \times d$ is used. 
Elements below and on the main diagonal are zero, while upper-diagonal entries $\boldsymbol{\Psi}_k(u,v)$ (where $u < v$) specify rotation angles for variable pairs. 
Algorithm~\ref{alg:RotationControlled} is used to construct $\mathbf{R}_k$ from $\boldsymbol{\Psi}_k$.  
The resulting $\mathbf{R}_k$ captures all specified pairwise dependencies while preserving orthogonality.

\begin{algorithm}[!tp]
\footnotesize
$\mathbf{R}_{k}\gets \mathbf{I}_d$\;
\For{$u=1$ \KwTo $d-1$}{
  \For{$v=u+1$ \KwTo $d$}{
    \If{$\boldsymbol{\Psi}_{k}(u,v)\neq 0$}{
      $\mathbf{R}_{k}\gets \mathbf{R}_{k}\,\mathbf{G}^{(u,v)}\!\bigl(\boldsymbol{\Psi}_{k}(u,v)\bigr)$ \tcp*{See Eq.~\eqref{eq:givens}}
    }
  }
}
\caption{Constructing $\mathbf{R}_k$ from angle matrix $\boldsymbol{\Psi}_k$ via plane rotations.}
\label{alg:RotationControlled}
\end{algorithm}

This sequential multiplication ensures that the final $\mathbf{R}_k$ integrates all specified pairwise interactions consistently while maintaining orthogonality.  
Note that the order of multiplications (outer loop over $u$, inner loop over $v$) defines a deterministic construction scheme. 
Different orders would yield alternative orthogonal matrices that encode equivalent interaction patterns.

\subsection{Transformations for Local Optima and Ruggedness}
\label{sec:transforms}

Let $\mathbf{a}=\mathbf{R}_k(\mathbf{x}-\mathbf{c}_k)\in\mathbb{R}^d$ denote the pre-transform internal coordinates of component $k$ (cf. Equation~\eqref{eq:InternalCoordinate}). A transformation $\mathfrak{T}_k:\mathbb{R}^d\!\to\!\mathbb{R}^d$ warps $\mathbf{a}$ before it enters the component function, injecting controlled local optima and fine-grained ruggedness. 

All transformations are designed to be \emph{center- and curvature-preserving}: $\mathfrak{T}_k(\mathbf{0})=\mathbf{0}$ and $\nabla \mathfrak{T}_k(\mathbf{0})=\mathbf{I}_d$ where $\nabla\mathfrak{T}_k$ denotes the Jacobian (matrix of first derivatives). This ensures that the component center remains a stationary point and the local quadratic structure is unchanged. 
These constraints isolate ruggedness to higher-order effects while keeping parameter semantics stable: the scaling factors $\kappa_{k,i}^{\pm}$ continue to modulate anisotropy and conditioning, while the exponent $\mathfrak{p}_k$ continues to control basin curvature. This design also supports systematic comparability across different configurations.

When $\mathfrak{T}_k$ is the identity map, the basin remains smooth. 
We group transformations into two classes:
\begin{itemize}
  \item \emph{Element-wise (separable, interaction-preserving).}  
  These operators act independently on each coordinate and therefore preserve the existing variable-interaction structure of the baseline.  
  PORTAL currently includes three element-wise operators: additive periodic perturbation, log-sinusoidal phase modulation, and wavelet-inspired modulation.  

  \item \emph{Coupling (non-separable, interaction-inducing).}  
  These operators depend jointly on multiple coordinates, introducing explicit cross-coordinate coupling even when the baseline function is otherwise separable.  
  PORTAL currently includes two coupling operators: tensor interference and radial polynomial–trigonometric hybrid.  
\end{itemize}

While all element-wise operators produce axis-aligned ruggedness patterns, they differ significantly in their spatial characteristics and intensity profiles. Additive periodic perturbation creates regularly-spaced ripples with approximately uniform amplitude once the envelope saturates, maintaining consistent feature size across the basin. 
Log-sinusoidal modulation produces self-similar patterns where both feature spacing and relative intensity scale logarithmically with distance, meaning that perturbations become more compressed and relatively weaker near the center and thus create a fractal-like structure.
Wavelet-inspired modulation creates features localized to a specific radial band via the Gaussian envelope, with the strongest perturbations concentrated in an intermediate zone and decay both toward the center and at large radii. 
The two coupling operators are fundamentally distinct from each other and from all element-wise operators: tensor interference creates rectilinear grid patterns through multiplicative gating across coordinates, while the radial operator produces rotationally-symmetric concentric rings without directional preference.

Beyond these two families, PORTAL also supports \emph{compositional use}: multiple transformations can be applied in sequence to the same component, producing hybrid ruggedness patterns. Because these operators are non-commutative, the choice and order of composition significantly expands the diversity of basin structures. 

For each operator below we give its explicit form, parameter roles (e.g., amplitudes, frequencies, envelopes), and qualitative impact on the pattern of local optima (symmetry, spacing, depth, and extent). A small constant $\varepsilon>0$ is used where needed for numerical stability.

%%%%%%%%%%%%%%%%%%%%%%%%%%%%%%%%%%%%%%%%%%%%%%%%%%%%%%%%%%%%%%%%%%%%%%%%%%%%%%%
\subsubsection{Element-Wise Transformations (Separable, Interaction-Preserving)}
\label{sec:sec:sec:transformationsElement}

\paragraph{Additive Periodic Perturbation}

This transformation adds sinusoidal ripples along each coordinate axis, modulated by a saturating exponential envelope $\,1-e^{-\gamma|a|}\,$, applied element-wise to the internal coordinates. 
For coordinate $a_{k,i}$, with direction-specific parameters $(\mu_{k,i}^{\pm},\gamma_{k,i}^{\pm},\omega_{k,i}^{\pm})$, the transformation is defined as
\begin{align}
\label{eq:additive_periodic}
\resizebox{0.98\columnwidth}{!}{$
\bigl[\mathfrak{T}_k(\mathbf{a})\bigr]_i
=
a_{k,i}
+
\mu_{k,i}^{\,\mathrm{sgn}(a_{k,i})}\,
a_{k,i}\!\left(1-e^{-\gamma_{k,i}^{\,\mathrm{sgn}(a_{k,i})}|a_{k,i}|}\right)
\sin\!\bigl(\omega_{k,i}^{\,\mathrm{sgn}(a_{k,i})}|a_{k,i}|\bigr),$}
\end{align}
where $\mathrm{sgn}(\cdot)\in\{-,+\}$ selects the parameter triplet for the negative or positive half-axis, respectively.
The three parameters have intuitive roles: $\mu_{k,i}^{\pm}\!\ge 0$ sets ripple strength (depth and height of induced extrema), $\gamma_{k,i}^{\pm}\!>\!0$ controls how quickly the envelope ramps up from the center, and $\omega_{k,i}^{\pm}\!>\!0$ sets the spacing of successive extrema (zeros near $|a_{k,i}| \approx m\pi/\omega_{k,i}^{\pm}$ and local maxima/minima near $|a_{k,i}| \approx (2m{+}1)\pi/(2\omega_{k,i}^{\pm})$ along axis $i$). The envelope factor $(1-e^{-\gamma|a|})$ starts at $0$ near the center and saturates to $1$ as $|a|$ grows, initially suppressing perturbations near the origin and then allowing the sinusoidal modulation to reach full effect.

Near the center, a Taylor expansion shows the additive term is $O(|a_{k,i}|^3)$ as $a_{k,i}\!\to\!0$, ensuring that $\mathfrak{T}_k(\mathbf{0})=\mathbf{0}$ and $\nabla\mathfrak{T}_k(\mathbf{0})=\mathbf{I}_d$ hold. This preserves the component center and local quadratic structure while injecting only higher-order ruggedness.

Because the sinusoid and exponential envelope are even in $|a_{k,i}|$ while the factor $a_{k,i}$ is odd, the perturbation exhibits odd parity in $a_{k,i}$. When the positive and negative parameter triplets are matched ($\mu^+=\mu^-$, $\gamma^+=\gamma^-$, $\omega^+=\omega^-$), the overall map becomes odd: $\mathfrak{T}_k(-\mathbf{a})=-\mathfrak{T}_k(\mathbf{a})$, yielding sign-symmetric ripples about the origin.

Acting coordinate-wise, the operator does not introduce cross-coordinate coupling. Asymmetries across axes or directions can be introduced deliberately by choosing different $(\mu_{k,i}^{\pm},\gamma_{k,i}^{\pm},\omega_{k,i}^{\pm})$ values. The resulting landscape exhibits axis-aligned ripple trains that are suppressed near the center by the envelope and, once the envelope has saturated, attain their full sinusoidal modulation; the perturbation's envelope then grows roughly linearly with $|a_{k,i}|$ due to the prefactor $a_{k,i}$.
An example of this transformation with asymmetric parameter settings is illustrated in Fig.~\ref{fig:add-per}.

\paragraph{Log-sinusoidal phase modulation}
This element-wise transformation modulates each coordinate by a sinusoidal signal in \emph{log-space}, producing scale-sensitive, self-similar irregularities: patterns repeat across decades of $|a_{k,i}|$ rather than at fixed linear spacings. With direction-specific parameters $\mu_{k,i}^{\pm}\!\ge 0$, two log-space frequencies $\omega_{1,i}^{\pm},\omega_{2,i}^{\pm}\!>\!0$, and a small numerical constant $\varepsilon=10^{-12}$, the action on coordinate $a_{k,i}$ is

\begin{align}
\label{eq:logsinusoidal}
\bigl[\mathfrak{T}_k(\mathbf{a})\bigr]_i
&= \mathrm{sgn}(a_{k,i})\,\exp\Bigl(\log(|a_{k,i}|+\varepsilon) \nonumber\\
&\quad +\mu_{k,i}^{\,\mathrm{sgn}(a_{k,i})}\bigl[\sin(\omega_{1,i}^{\,\mathrm{sgn}(a_{k,i})}\log(|a_{k,i}|+\varepsilon)) \nonumber\\
&\quad\quad +\sin(\omega_{2,i}^{\,\mathrm{sgn}(a_{k,i})}\log(|a_{k,i}|+\varepsilon))\bigr]\Bigr),
\end{align}
where $\mathrm{sgn}(\cdot)\in\{-,+\}$ selects the direction-specific parameter triplet.
The amplitude $\mu_{k,i}^{\pm}$ controls modulation strength (larger values increase deviation from the identity and deepen induced extrema),
while the log-frequencies $\omega_{1,i}^{\pm},\omega_{2,i}^{\pm}$ determine how many oscillations occur per multiplicative change in scale. With a single frequency $\omega$, salient features appear at geometrically spaced radii $|a_{k,i}|\approx \exp(m\pi/\omega)$ for integers $m$; using two frequencies produces multi-scale interference (beating), enriching the ruggedness pattern.

The regularization with $\varepsilon$ ensures smoothness at the origin: the logarithmic argument is regularized, so both the log and sinusoidal terms remain finite as $a \to 0$. Expanding in powers of $a$ shows that deviations from the identity are $O(\mu|a|)$, so the Jacobian at the origin is the identity, i.e., $\mathfrak{T}_k(\mathbf{0})=\mathbf{0}$ and $\nabla\mathfrak{T}_k(\mathbf{0})=\mathbf{I}_d$, preserving the component center and its local quadratic structure. For $|a_{k,i}|\gg\varepsilon$, the induced irregularity varies with $\log|a_{k,i}|$ rather than growing unboundedly with $|a_{k,i}|$. When the positive and negative parameter triplets are matched ($\mu^+=\mu^-$, $\omega_{1}^+=\omega_{1}^-$, $\omega_{2}^+=\omega_{2}^-$), the transformation is odd, $\bigl[\mathfrak{T}_k(-\mathbf{a})\bigr]_i=-\bigl[\mathfrak{T}_k(\mathbf{a})\bigr]_i$, yielding sign-symmetric perturbations about the origin. Because the operator acts coordinate-wise, it preserves the existing variable--interaction structure; asymmetries across axes or directions can be introduced deliberately by choosing different $(\mu_{k,i}^{\pm},\omega_{1,i}^{\pm},\omega_{2,i}^{\pm})$.
A representative visualization of the log-sinusoidal effects under asymmetric settings is shown in Figure~\ref{fig:log-sin}.

\paragraph{Wavelet-Inspired Modulation}

This element-wise transformation  (using a Gaussian-windowed sinusoidal carrier with odd symmetry) creates symmetric pairs of local optima along each coordinate axis while preserving the component center and its first-order basin shape. For coordinate $a_{k,i}$, let $t=|a_{k,i}|$ and select direction-specific parameters $(\mu_{k,i}^{\pm},\omega_{k,i}^{\pm},\ell_{k,i}^{\pm})$ with amplitudes $\mu_{k,i}^{\pm}\!\ge 0$, frequencies $\omega_{k,i}^{\pm}\!>\!0$, and envelope lengths $\ell_{k,i}^{\pm}\!>\!0$. By default, we tie spatial extent to frequency via a component-level extent scale $\eta_k\!>\!0$:
\[
\ell_{k,i}^{\pm} \;=\; \frac{\eta_k}{\omega_{k,i}^{\pm}}.
\]

The transformation uses an even envelope–carrier function:
\begin{align}
\resizebox{0.98\columnwidth}{!}{$
\phi_{k,i}^{\,\mathrm{sgn}(a_{k,i})}(t)
= \Bigl(\tfrac{t}{\ell_{k,i}^{\,\mathrm{sgn}(a_{k,i})}}\Bigr)^{\!2}
  \exp\!\Bigl(-\tfrac{t^2}{\bigl(\ell_{k,i}^{\,\mathrm{sgn}(a_{k,i})}\bigr)^2}\Bigr)\,
  \sin\!\Bigl(\omega_{k,i}^{\,\mathrm{sgn}(a_{k,i})} t - \tfrac{\pi}{2}\Bigr),\nonumber$}
\end{align}

applied through an odd, center-preserving lift:
\begin{align}
\label{eq:wavelet_transform}
\bigl[\mathfrak{T}_k(\mathbf{a})\bigr]_i
=
a_{k,i}
\;+\;
\mu_{k,i}^{\,\mathrm{sgn}(a_{k,i})}\;
\phi_{k,i}^{\,\mathrm{sgn}(a_{k,i})}(|a_{k,i}|)\;
\frac{a_{k,i}}{|a_{k,i}|+\varepsilon},
\end{align}
where $\varepsilon=10^{-12}$.

The factor $a_{k,i}/(|a_{k,i}|+\varepsilon)$ transmits the sign and keeps the perturbation bounded at the origin. Since $\phi_{k,i}^{\,\mathrm{sgn}(a_{k,i})}(\cdot)$ is even in $t$ and the sign factor is odd in $a_{k,i}$, the additive term is odd. A local expansion gives $\phi(t)= -\bigl(t/\ell\bigr)^{2}+O(t^{4})$ as $t\!\to\!0$, so the additive term in \eqref{eq:wavelet_transform} scales like $O(|a_{k,i}|^{2})$. Consequently, $\mathfrak{T}_k(\mathbf{0})=\mathbf{0}$ and $\nabla\mathfrak{T}_k(\mathbf{0})=\mathbf{I}_d$, preserving the component center and first-order behavior while injecting higher-order ruggedness.

The parameters control interpretable aspects of the pattern. The amplitude $\mu_{k,i}^{\pm}$ sets the depth/height of extrema. The frequency $\omega_{k,i}^{\pm}$ governs spacing along axis $i$ (features near $t \approx \frac{(2m+1)\pi}{2\,\omega_{k,i}^{\pm}}$, spacing $\approx \pi/\omega_{k,i}^{\pm}$).
The envelope length $\ell_{k,i}^{\pm}$ limits spatial extent; larger $\ell$ spreads oscillations farther from the center, while smaller $\ell$ concentrates them.
Using $\ell_{k,i}^{\pm}=\eta_k/\omega_{k,i}^{\pm}$ ties extent to frequency via $\eta_k$, letting the user to vary spacing without unintentionally changing range. Because the operator is coordinate-wise, it introduces no cross-coordinate coupling; symmetry can be kept (matched $+/-$ triplets) or intentionally broken by choosing different $(\mu,\omega,\ell)$ per direction.
Typically, the user specifies the per-direction frequency $\omega_{k,i}^{\pm}$ and a single component-level extent scale $\eta_k>0$. The envelope length is then \emph{derived} by the default convention $\ell_{k,i}^{\pm}=\eta_k/\omega_{k,i}^{\pm}$. This keeps roles decoupled and intuitive: $\omega_{k,i}^{\pm}$ sets spacing, $\eta_k$ sets overall extent, and the approximate number of visible lobes $\omega_{k,i}^{\pm}\ell_{k,i}^{\pm}/\pi\approx \eta_k/\pi$ stays roughly constant when adjusting $\omega_{k,i}^{\pm}$. 
An illustrative example of this transformation is shown in Figure~\ref{fig:wavelet}.

\subsubsection{Coupling Transformations (Non-Separable, Interaction-Inducing)}
\label{sec:transforms-coupling}

\paragraph{Tensor Interference}
This transformation injects explicit cross-coordinate coupling by modulating each coordinate with an odd sinusoid in that coordinate, gated by an \emph{even-in-the-others} interference term. Let $\mathbf{a}=\mathbf{R}_k(\mathbf{x}-\mathbf{c}_k)\in\mathbb{R}^d$ denote the internal coordinates. For each output coordinate $i$, define
\[
\Phi_k^{(i)}(\mathbf{a})
\;=\;
\prod_{j\neq i}\sin^2\!\bigl(\omega_{k,j}^{\,\mathrm{sgn}(a_{k,j})}\,a_{k,j}\bigr),
\]
where $\omega_{k,j}^{\pm}>0$, and apply
\begin{align}
\label{eq:tensor_interference}
\bigl[\mathfrak{T}_k(\mathbf{a})\bigr]_i
\;=\;
a_{k,i}
\;+\;
\mu_{k,i}^{\,\mathrm{sgn}(a_{k,i})}\,
\sin\!\bigl(\omega_{k,i}^{\,\mathrm{sgn}(a_{k,i})}\,a_{k,i}\bigr)\,
\Phi_k^{(i)}(\mathbf{a}),
\end{align}
with $\mu_{k,i}^{\pm}\ge 0$.
The multiplier $\Phi_k^{(i)}$ is even in all $a_{k,j}$ with $j\neq i$, so it acts as a gate: it activates perturbations near constructive-interference bands ($a_{k,j}\!\approx\!(2m{+}1)\pi/(2\omega_{k,j}^{\pm})$) and suppresses them near nodes ($a_{k,j}\!\approx\!m\pi/\omega_{k,j}^{\pm}$). The resulting landscapes show checkerboard patterns for $d=2$ or hyper--checkerboard patterns for $d>2$, with spacing $\pi/\omega_{k,j}^{\pm}$ along axis $j$ and perturbation strength along $i$ governed by $\mu_{k,i}^{\pm}$.

A key consideration is dimensional scaling. Since $\mathbb{E}[\sin^2(\cdot)]=\tfrac12$, the expected gate magnitude satisfies $\mathbb{E}[\Phi_k^{(i)}]\approx 2^{-(d-1)}$. Thus, without correction, the perturbation vanishes exponentially as $d$ grows. To counteract this, the amplitude is scaled as
\[
\mu_{k,i}^{\pm} = \mu_{0,k,i}^{\pm}\,2^{d-1},
\]
where $\mu_{0,k,i}^{\pm}$ is a dimension-independent base amplitude. With this normalization, $\mu_{0,k,i}^{\pm}$ consistently controls perturbation strength regardless of dimensionality.

Near the origin, $\sin(\omega a_{k,i})=\omega a_{k,i}+O(a_{k,i}^3)$ and $\sin^2(\omega a_{k,j})=O(a_{k,j}^2)$ for $j\neq i$, so the additive term scales like
\[
O\!\Bigl(a_{k,i}\,\prod_{j\neq i} a_{k,j}^{2}\Bigr).
\]
This guarantees $\mathfrak{T}_k(\mathbf{0})=\mathbf{0}$ and $\nabla\mathfrak{T}_k(\mathbf{0})=\mathbf{I}_d$, preserving the component center and its local quadratic structure while introducing only higher-order interactions. Because the sinusoid in $a_{k,i}$ is odd and the multiplier is even in the other coordinates, the perturbation to coordinate $i$ is odd in $a_{k,i}$. When parameters match across directions ($\mu_{k,i}^+=\mu_{k,i}^-$ and $\omega_{k,i}^+=\omega_{k,i}^-$), the full map is odd: $\mathfrak{T}_k(-\mathbf{a})=-\mathfrak{T}_k(\mathbf{a})$, producing sign-symmetric interference.

Unlike the element-wise operators, tensor interference \emph{explicitly} induces dense cross-coordinate coupling: each output coordinate depends on all others through $\Phi_k^{(i)}$. While the product can be restricted to a subset of coordinates to localize interactions, PORTAL uses full gating by default since variable interaction structure is primarily controlled through the rotation and block-wise mechanisms.
A two-dimensional illustration of the tensor interference transformation under asymmetric parameter settings is shown in Figure~\ref{fig:tensor}.

\paragraph{Radial polynomial–trigonometric hybrid}

This transformation introduces isotropic ruggedness by modulating the vector $\mathbf{a}$ with a radius-dependent carrier, creating concentric rings of local optima that depend only on $r=\|\mathbf{a}\|_2$. With component-level parameters $\mu_k\!\ge 0$, $0<p_k<1$, $q_k>0$, and $\omega_k>0$, define
\[
\phi_k(r)=r^{p_k}\,\sin\!\bigl(\omega_k\,r^{q_k}\bigr).
\]
We use the following center- and first-order-preserving additive form:
\begin{align}
\label{eq:radial_hybrid_add}
\mathfrak{T}_k(\mathbf{a})
= \mathbf{a} \;+\; \mu_k\,\phi_k(r)\,\frac{\mathbf{a}}{r+\varepsilon},
\qquad \varepsilon=10^{-12}.
\end{align}
Near the origin, $\sin(\omega_k r^{q_k})=\omega_k r^{q_k}+O(r^{3q_k})$, hence $\phi_k(r)=\omega_k r^{p_k+q_k}+O(r^{p_k+3q_k})$, and
$\mathfrak{T}_k(\mathbf{0})=\mathbf{0}$ and $\nabla\mathfrak{T}_k(\mathbf{0})=\mathbf{I}_d$ hold. The map is odd and rotation-equivariant; because $\phi_k$ depends only on $r$, the transform is non-separable (it induces cross-coordinate coupling) yet introduces no directional bias. The amplitude $\mu_k$ sets the modulation strength; $p_k$ controls how the magnitude ramps with radius, and $q_k$ controls the radial spacing via $\omega_k r^{q_k}$ (uniform for $q_k\!=\!1$, widening for $0\!<\!q_k\!<\!1$, compressing for $q_k\!>\!1$).
Figure~\ref{fig:radial} illustrates the effect of this radial hybrid operator under asymmetric parameter settings.

\subsubsection{Transformation Composition}

While element-wise and coupling transformations each enrich landscape characteristics in distinct ways, they can also be \emph{hybridized} by applying multiple transformations in sequence to the same component. 
PORTAL allows users to compose transformations sequentially, creating hybrid ruggedness patterns that combine characteristics from different operator types. The number, type, and order of transformations are user-specified. 

For component $k$, a sequence $\mathfrak{T}_k^{(1)}, \mathfrak{T}_k^{(2)}, \ldots, \mathfrak{T}_k^{(n)}$ is applied successively to the internal coordinates:
\begin{align}
\mathbf{z}_k = \mathfrak{T}_k^{(n)} \circ \mathfrak{T}_k^{(n-1)} \circ \cdots \circ \mathfrak{T}_k^{(2)} \circ \mathfrak{T}_k^{(1)}\bigl(\mathbf{R}_k(\mathbf{x}-\mathbf{c}_k)\bigr),
\end{align}
where each $\mathfrak{T}_k^{(i)}$ may be chosen independently from the available element-wise or coupling operators. Since each transformation individually satisfies $\mathfrak{T}_k^{(i)}(\mathbf{0})=\mathbf{0}$ and $\nabla\mathfrak{T}_k^{(i)}(\mathbf{0})=\mathbf{I}_d$, their composition likewise preserves the component center and its local quadratic structure.

\begin{figure*}[!t]
\centering
\begin{tabular}{ccc}
  % Row 1 — curvature with isotropic kappa
   \subfigure[{\scriptsize Illustration of the first component function defined in~\eqref{eq:Form1}, showing the effects of direction-specific exponents $\mathfrak{p}_{k,i}^{\pm}$ and anisotropy factors $\kappa_{k,i}^{\pm}$ on curvature and conditioning.}]{%
    \includegraphics[width=0.3\textwidth]{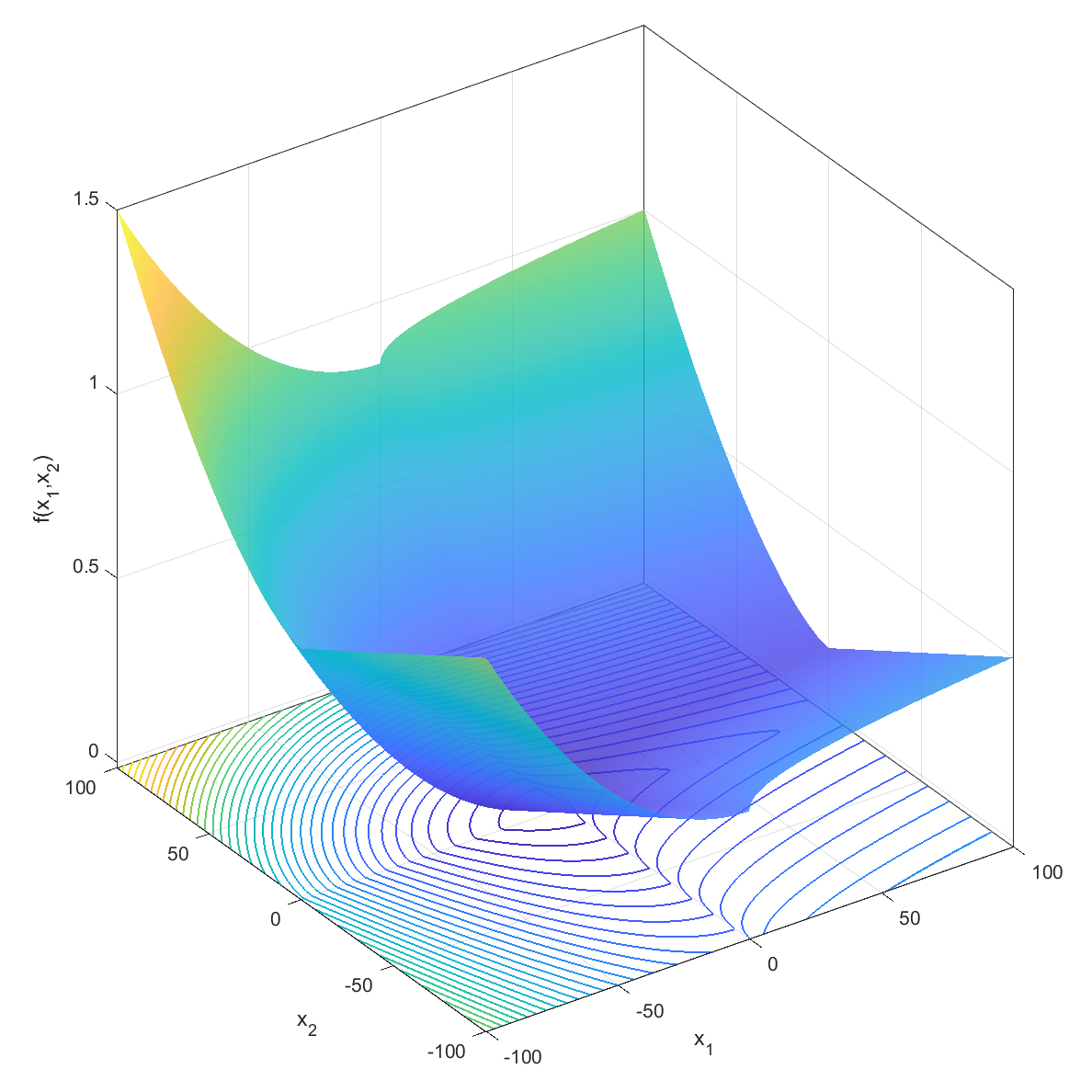}\label{fig:form1}} &

      \subfigure[{\scriptsize An example generated using the single-exponent component function in Equation~\eqref{eq:Form2}, shown under different anisotropy settings.}]{%
    \includegraphics[width=0.3\textwidth]{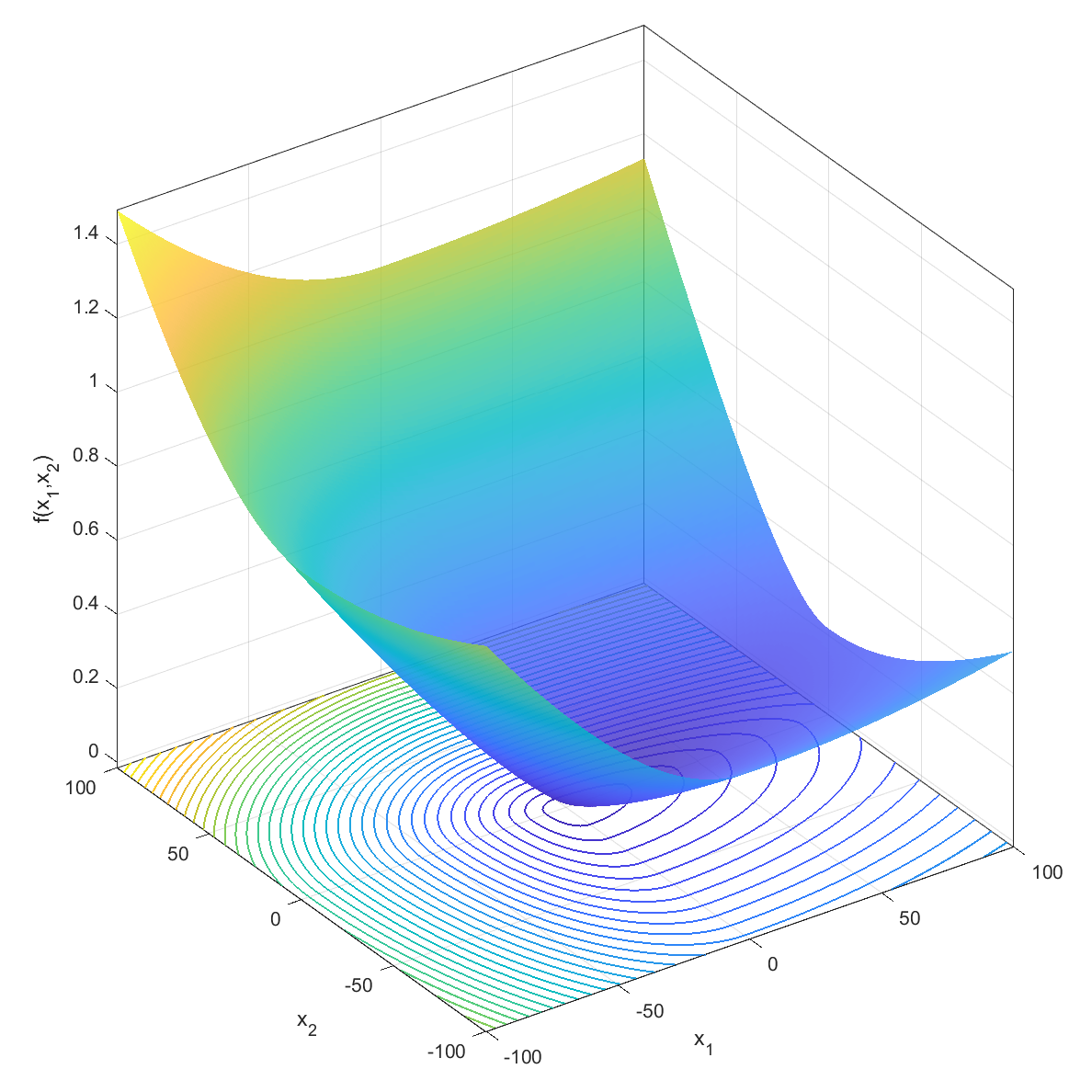}\label{fig:form2}} &

          \subfigure[{\scriptsize 
    Example of a two-component landscape constructed via the $\min$ operator using one Form~1 component and one Form~2 component, each with different exponents.}]{%
    \includegraphics[width=0.3\textwidth]{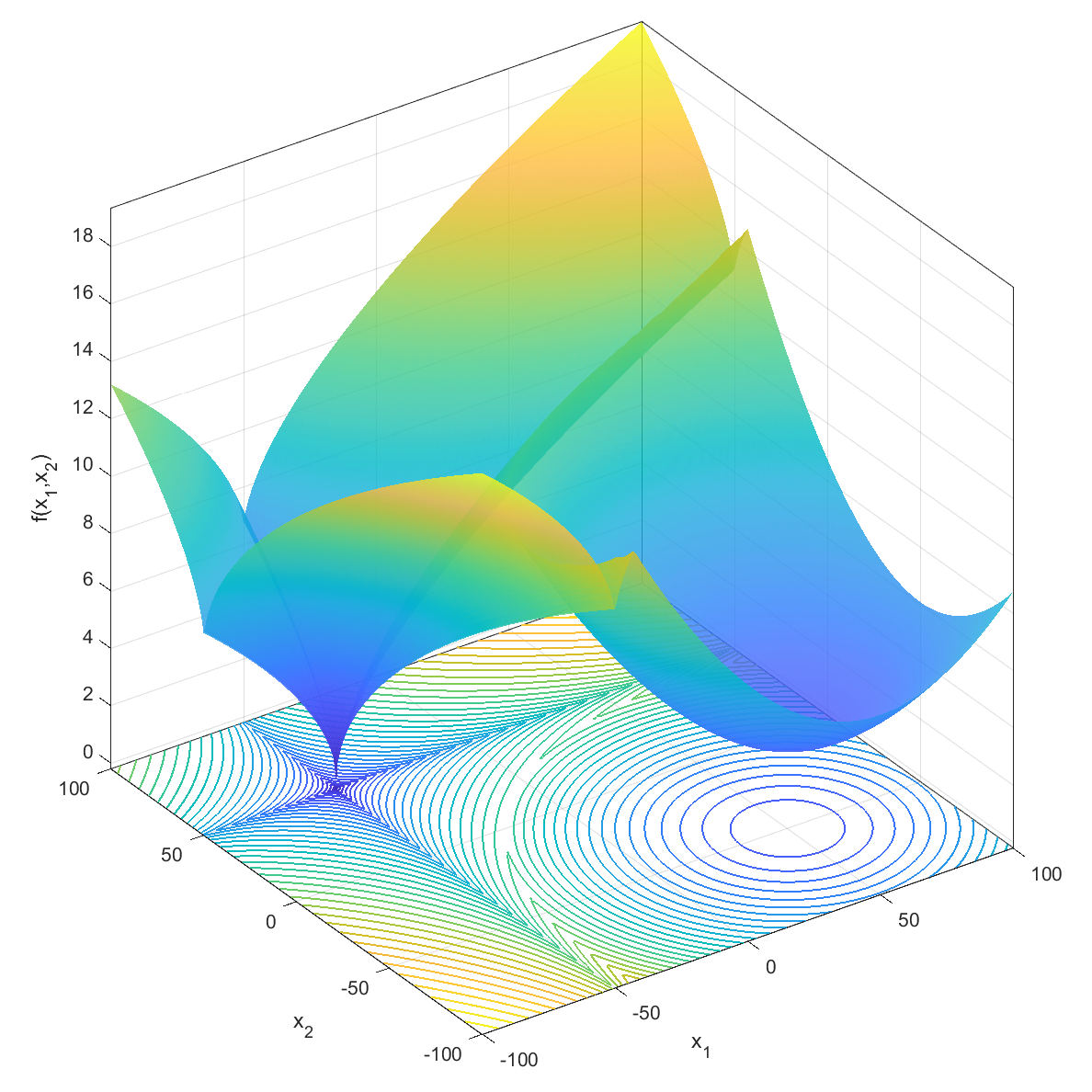}\label{fig:layer2}}
    \\
      \subfigure[{\scriptsize 
    Additive periodic perturbation (Eq.~\ref{eq:additive_periodic}) applied with asymmetric parameter settings, producing axis-aligned ripple patterns with envelope-controlled amplitude.}]{%
    \includegraphics[width=0.3\textwidth]{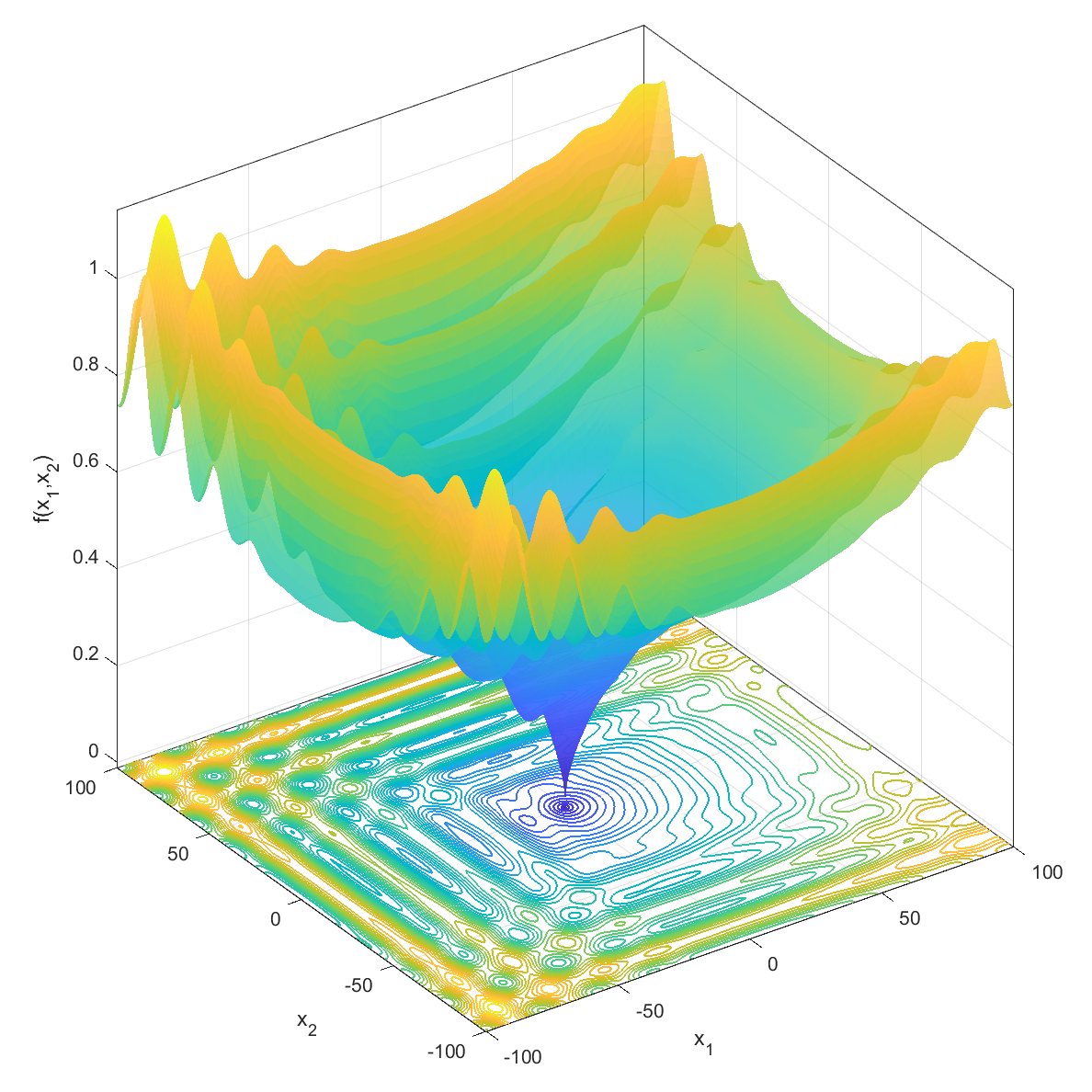}\label{fig:add-per}} &

     \subfigure[{\scriptsize 
    Log-sinusoidal phase modulation (Equation~\eqref{eq:logsinusoidal}) under asymmetric parameter settings, illustrating multi-scale, log-spaced ruggedness.}]{%
    \includegraphics[width=0.3\textwidth]{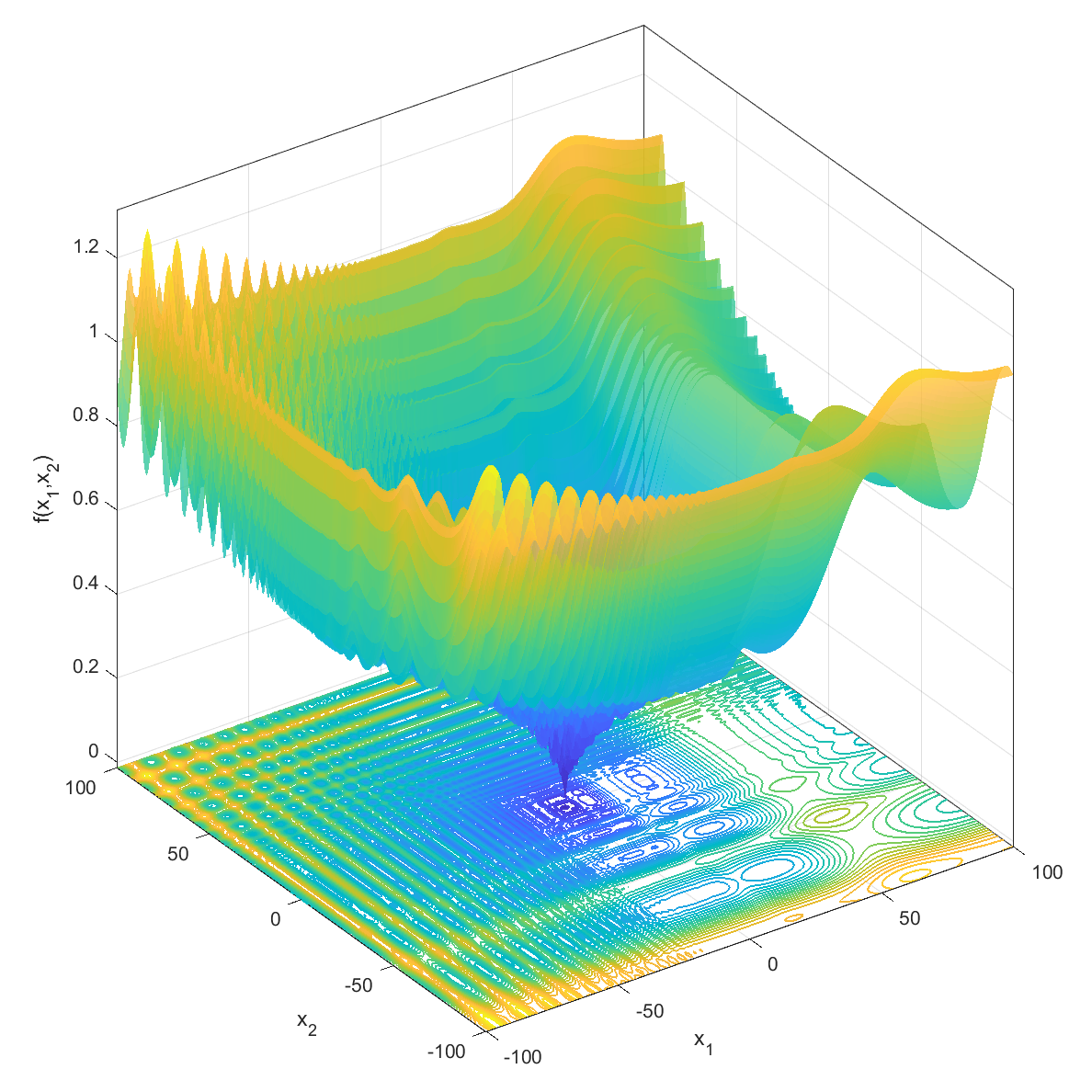}\label{fig:log-sin}} &

      \subfigure[{\scriptsize 
    Wavelet-inspired modulation (Equation~\eqref{eq:wavelet_transform}), showing how amplitude, frequency, and extent parameters control the spacing and number of local extrema under asymmetric settings.}]{%
    \includegraphics[width=0.3\textwidth]{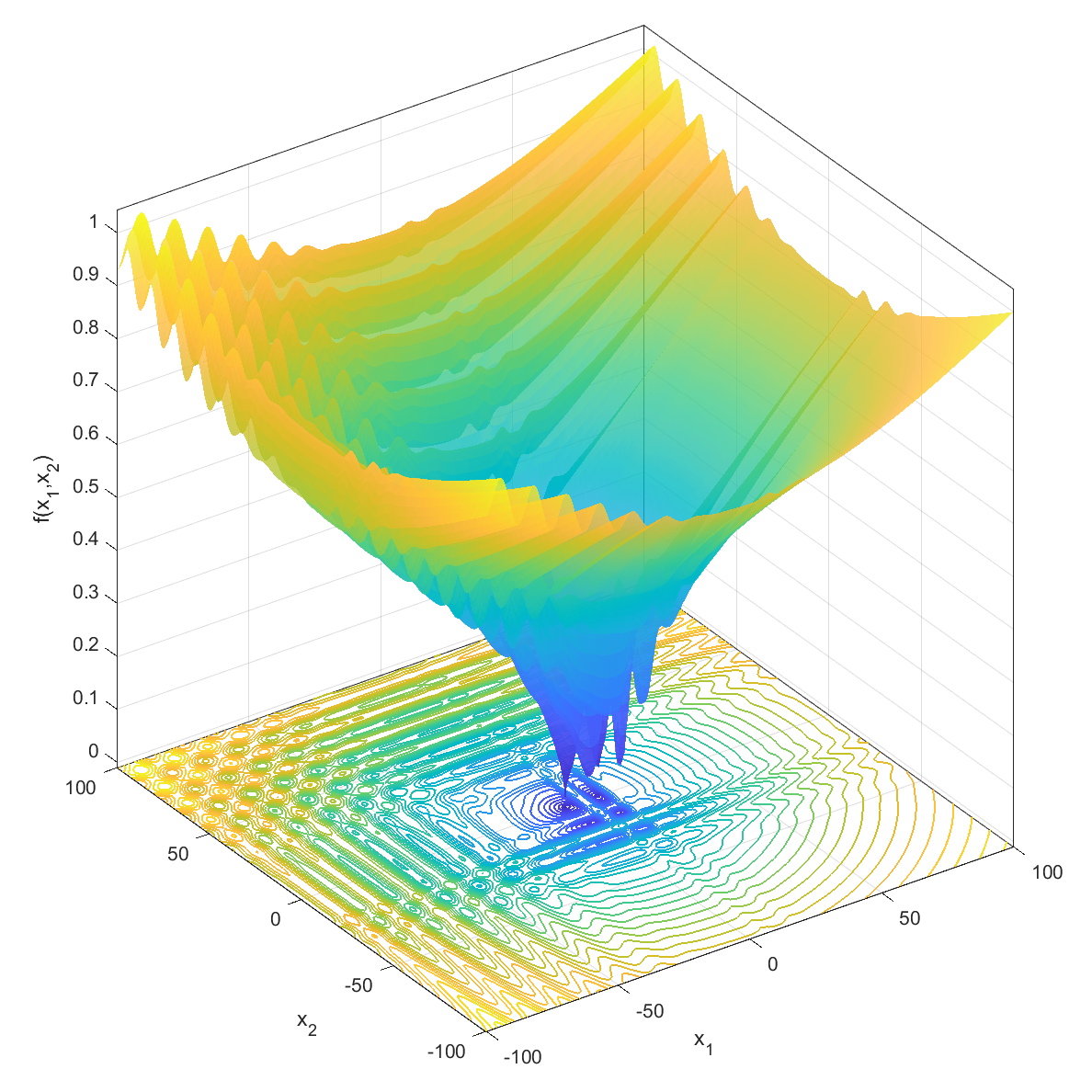}\label{fig:wavelet}} \\

      \subfigure[{\scriptsize 
    Tensor interference transformation (Equation~\eqref{eq:tensor_interference}) showing cross-coordinate coupling and interference-induced checkerboard patterns under asymmetric settings.}]{%
    \includegraphics[width=0.3\textwidth]{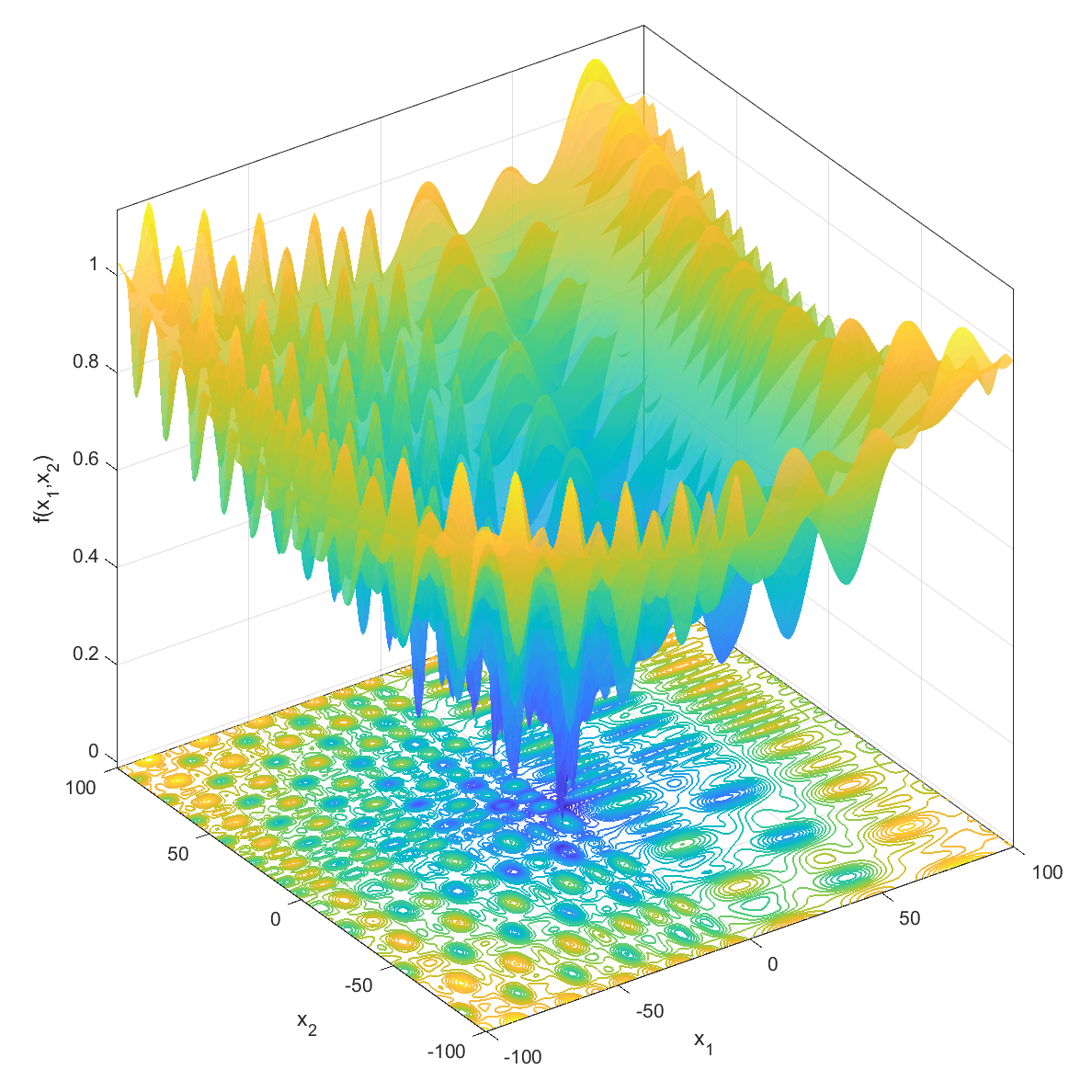}\label{fig:tensor}} &

      \subfigure[{\scriptsize 
    Radial polynomial--trigonometric transformation (Eq.~\eqref{eq:radial_hybrid_add}), creating concentric ring patterns through radius-dependent modulation.}]{%
    \includegraphics[width=0.3\textwidth]{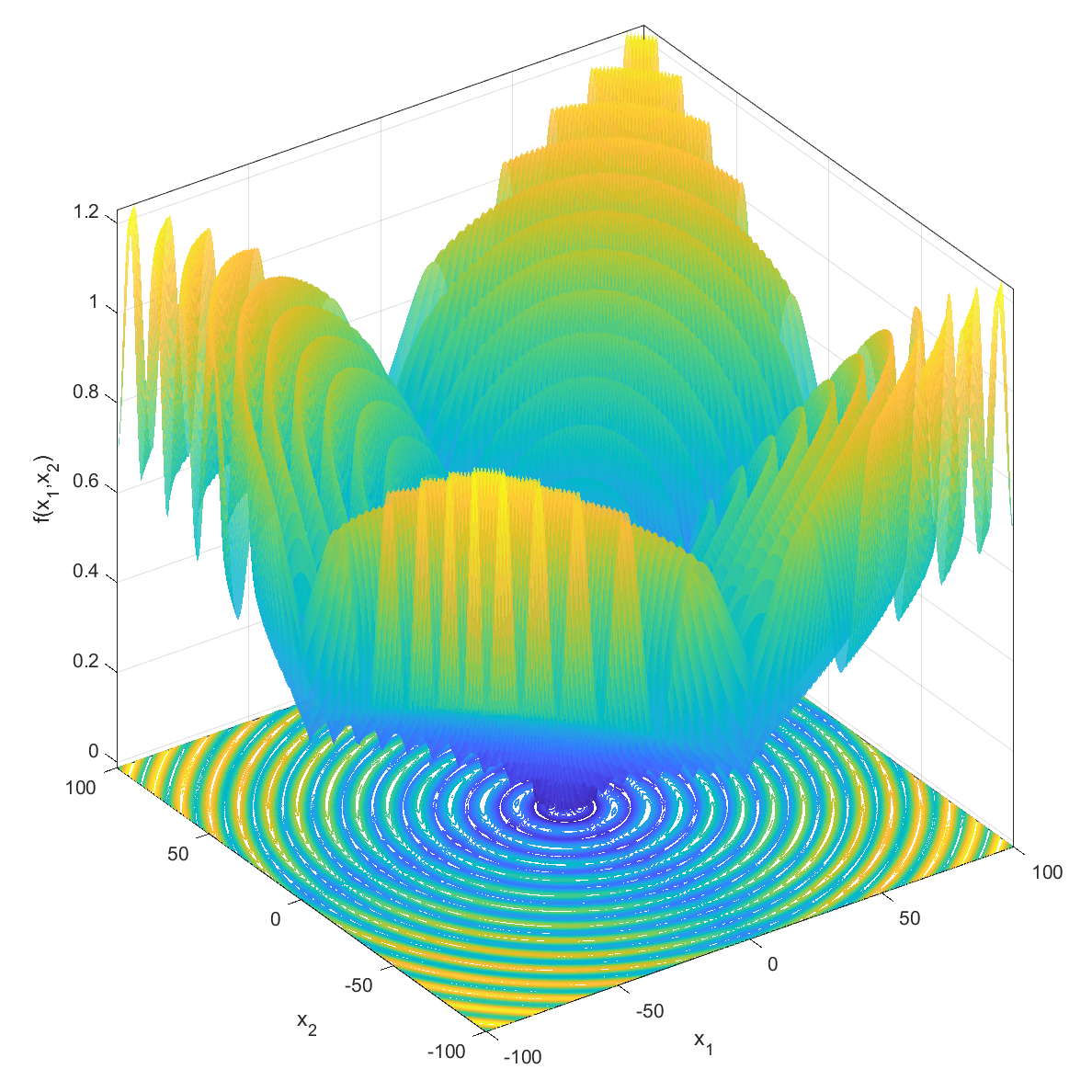}\label{fig:radial}} & 

      \subfigure[{\scriptsize 
    Hybrid transformation generated by sequential application of the Radial polynomial--trigonometric and tensor–interference transformations (see Equations~\eqref{eq:radial_hybrid_add} and~\eqref{eq:tensor_interference}).}]{%
    \includegraphics[width=0.3\textwidth]{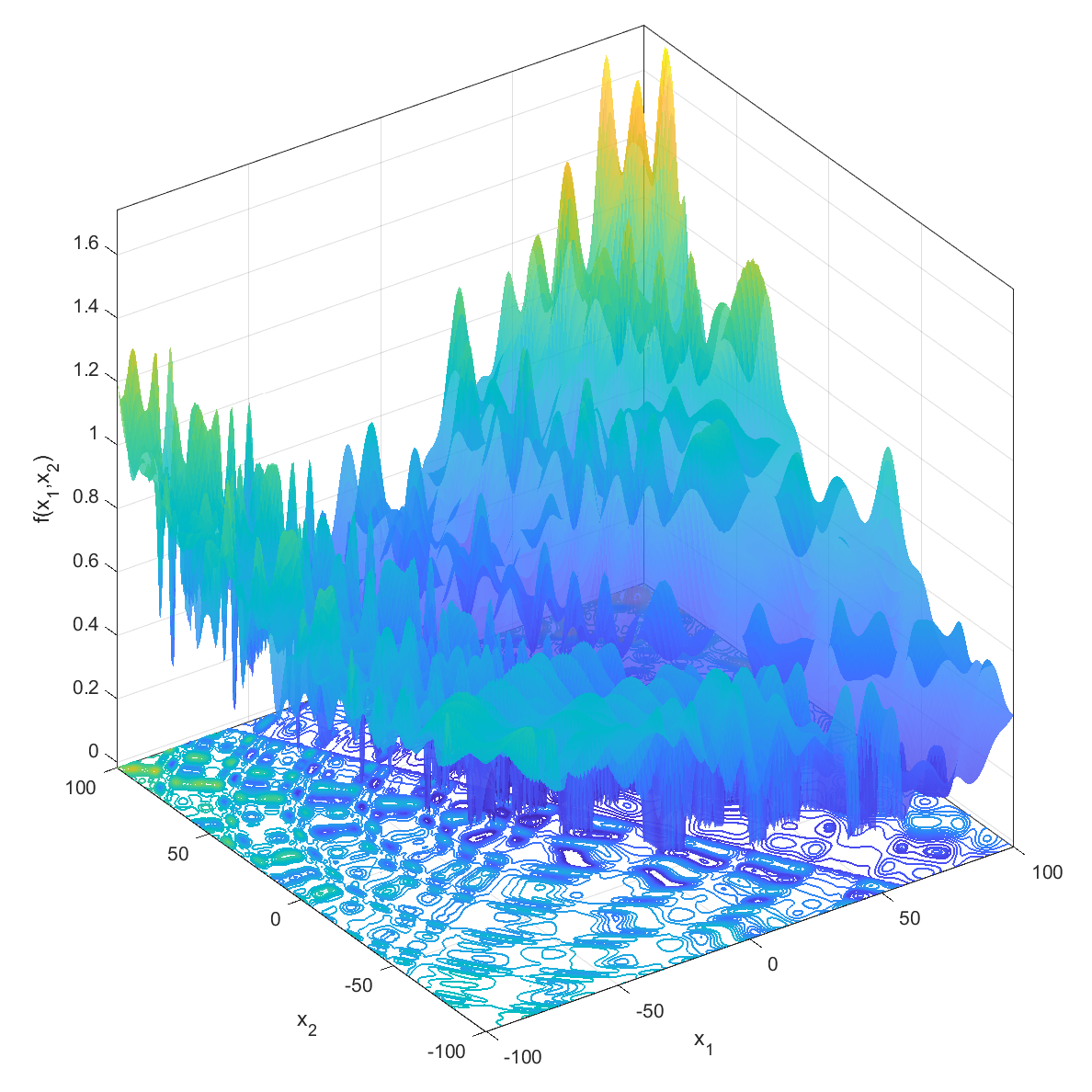}\label{fig:hybrid}} 

\end{tabular}
\caption{
Illustrative examples of PORTAL's component formulations, Layer~2 assembly (Layer 3 cannot be shown in 2-dimensional plots), and transformation operators. 
Subfigures~(a)--(b) show the two component function formulations (per-direction exponents and single-exponent form) with varied parameter settings. 
Subfigure~(c) displays a two-component landscape assembled via the Layer~2 operator. 
Subfigures~(d)--(h) demonstrate the effects of individual transformation operators, including additive periodic perturbation, log-sinusoidal modulation, wavelet-inspired modulation, tensor interference, and radial polynomial--trigonometric hybrid patterns, using asymmetric parameter settings. 
Subfigure~(i) presents a hybrid composition combining two transformations applied sequentially to a single component. 
}

\label{fig:Form1}
\end{figure*}

The compositional approach significantly enhances pattern diversity. Because transformations are generally non-commutative, changing their order can lead to qualitatively different basin patterns: $\mathfrak{T}_k^{(2)} \circ \mathfrak{T}_k^{(1)} \neq \mathfrak{T}_k^{(1)} \circ \mathfrak{T}_k^{(2)}$. Distinct transformation sequences thus yield ruggedness characteristics unattainable by any single operator. 
This compositional flexibility expands PORTAL's expressive capacity and enables the generation of richer, more varied instances for systematic evaluation and meta-algorithmic training.
An illustrative example of a hybrid transformation sequence, combining a Radial polynomial–trigonometric hybrid operator followed by tensor interference, is shown in Figure~\ref{fig:hybrid}.

% Fig 1 (Form 1; tuple notation per dimension & direction)
% =========================

\subsection{Summary of Parameters}
\label{sec:parameter_summary}

Table~\ref{tab:portal_parameters} provides a comprehensive reference of all PORTAL parameters, organized by framework layer and transformation type. 
For each parameter, the table specifies its mathematical symbol, description, cardinality (how many instances are needed), the equations in which it appears, suggested ranges based on empirical testing, and default values where applicable. 

Parameters marked as ``Computed'' are not free choices but are derived automatically from other user-specified values via the PORTAL's formulas. 
The suggested ranges reflect values that have been found to generate visually distinct and algorithmically challenging landscapes in typical usage, though users may explore beyond these ranges for specialized applications.

Cardinalities involving $d_j$ refer to the local dimensionality of the block containing the component (equal to the total dimension $d$ when using a single block). 
Transformation parameters are component-specific and optional: components may either use the identity transformation (no ruggedness) or compose multiple transformations as described in Section~\ref{sec:transforms}.

\begin{table*}
\centering
\scriptsize
\begin{threeparttable}
\caption{PORTAL Framework Parameters. Indexing conventions: $j$ for block, $k$ for component, $i$ for dimension, $\pm$ for direction (positive/negative).}
\label{tab:portal_parameters}
\renewcommand{\arraystretch}{1.25}
\setlength{\tabcolsep}{4pt}
\begin{tabular}{l l l l l c c}
\toprule
\textbf{Category} & \textbf{Symbol} & \textbf{Description} & \textbf{Cardinality} & \textbf{Equations} & \textbf{Suggested Range} & \textbf{Default}\\
\midrule

\multicolumn{7}{l}{\textbf{Layer 1: Component-Level Parameters (per component $k$)}} \\
\midrule
& $\mathbf{c}_k$ & Center position & $d_j$\tnote{\textbf{$\dagger$}} & \eqref{eq:InternalCoordinate} & Inside boundary & --- \\
& $\beta_k$ & Vertical offset at component center & $1$ & \eqref{eq:Form1}, \eqref{eq:Form2} & $\mathbb{R}$ & --- \\
& $\mathfrak{p}^{\pm}_{k,i}$ & Basin curvature factor for Form 1 & $2\times d_j$ (per sign) & \eqref{eq:Form1}, \eqref{eq:Form1Neutralization} & $(0.2, 1.2]$ & $0.6$ \\
& $\mathfrak{p}_k$ & Basin curvature factor for Form 2 & $1$ & \eqref{eq:Form2}, \eqref{eq:Form2Neutralization} & $(0.2, 1.2]$ & $0.6$ \\
& $\kappa^{\pm}_{k,i}$ & Anisotropy factors & $2\times d_j$ (per sign) & \eqref{eq:Form1}, \eqref{eq:Form2} & $>0$ & $100$ \\
& $\boldsymbol{\Psi}_k$ & Rotation angle matrix (upper-triangular) & $d_j\times d_j$ & \eqref{eq:givens} & $[-\pi,\pi)$ & $0$ (no rotation)\\
& $\mathbf{R}_k$ & Orthogonal rotation matrix & $d_j\times d_j$ & \eqref{eq:InternalCoordinate}, \eqref{eq:givens} & --- & Computed from $\boldsymbol{\Psi}_k$ \\
\cmidrule{2-7}
& \multicolumn{6}{l}{\textbf{Neutralization Parameters}}\\
\cmidrule{2-7}
& $\Delta_k$ & Scale factor--Target rise at reference radius & $1$ & \eqref{eq:Form1Neutralization}, \eqref{eq:Form2Neutralization} & $\ge 1$ & $100$ \\
& $r_{\mathrm{ref}_k}$ & Reference radius for neutralization & $1$ & \eqref{eq:Form1Neutralization}, \eqref{eq:Form2Neutralization} & $(0,100]$ & $100$ \\
& $\bar{\kappa}_{k,i}$ & Per-dimension average anisotropy & $d_j$ & \eqref{eq:Form2Neutralization} & --- & Computed from $\kappa_{k,i}^{\pm}$ \\
& $\bar{\kappa}_k$ & Component-level mean anisotropy & $1$ & \eqref{eq:Form1Neutralization} & --- & Computed from $\bar{\kappa}_{k,i}$ \\
& $\rho^{\pm}_{k,i}$ &  Neutralization factor for Form 1 & $2\times d_j$ (per sign) & \eqref{eq:Form1Neutralization} & --- & Computed \\
& $\rho_k$ & Neutralization factor for Form 2 & $1$ & \eqref{eq:Form2Neutralization} & --- & Computed \\
\midrule

\multicolumn{7}{l}{\textbf{Layer 2: Multi-Component Landscape}} \\
\midrule
& $|\mathcal{C}|$ & Number of components & scalar & \eqref{eq:Layer2Baseline} & $\in\{1,2,\ldots,25\}$ & --- \\
\midrule

\multicolumn{7}{l}{\textbf{Layer 3: Block Composition (per block $j$)}} \\
\midrule
& $b$ & Number of blocks & scalar & \eqref{eq:layer3} & $\ge 1$ & $1$ \\
& $d_j$ & Local dimension of block $j$ & scalar & --- & $\ge 2$ & --- \\
& $\mathcal{I}_j$ & Variable indices for block $j$ & set in $\{1,\ldots,d\}$ & \eqref{eq:layer3} & valid subsets & $\{1,\ldots,d\}$ \\
& $d$ & Total dimension & scalar & --- & --- & $\sum_{j}d_j$ \\
& $w_j$ & Weight of block $j$ & $1$ per block & \eqref{eq:layer3} & $(0,\infty)$ & $1$ \\
& $\Omega_j$ & Boundary domain for block $j$ & $d_j \times 2$ & --- & $[-100,100]$ & $[-100,100]$ \\
\midrule

\multicolumn{7}{l}{\textbf{Transformations (per component $k$)}} \\
\midrule
&\multicolumn{6}{l}{\textbf{Additive Periodic (Element-wise)}} \\
\cmidrule{2-7}
& $\mu^{\pm}_{k,i}$ & Ripple strength amplitude & $2\times d_j$ (per sign) & \eqref{eq:additive_periodic} & $[0.1, 0.7]$ & $0.4$ \\
& $\gamma^{\pm}_{k,i}$ & Saturating envelope rate & $2\times d_j$ (per sign) & \eqref{eq:additive_periodic} & $[0.002, 0.2]$\tnote{\textbf{$\ddagger$}} & $0.05$ \\
& $\omega^{\pm}_{k,i}$ & Oscillation frequency & $2\times d_j$ (per sign) & \eqref{eq:additive_periodic} & $[0.05, 1]$ & $1.0$ \\
\midrule
&\multicolumn{6}{l}{\textbf{Log-Sinusoidal (Element-wise)}} \\
\cmidrule{2-7}
& $\mu^{\pm}_{k,i}$ & Modulation strength & $2\times d_j$ (per sign) & \eqref{eq:logsinusoidal} & $[0.05, 0.50]$ & $0.25$ \\
& $\omega^{\pm}_{1,i}$ & Primary log-space frequency & $2\times d_j$ (per sign) & \eqref{eq:logsinusoidal} & $[5, 50]$ & $50$ \\
& $\omega^{\pm}_{2,i}$ & Secondary log-space frequency & $2\times d_j$ (per sign) & \eqref{eq:logsinusoidal} & $[5, 50]$ & $50$ \\
\midrule
&\multicolumn{6}{l}{\textbf{Wavelet-Inspired Modulation (Element-wise)}} \\
\cmidrule{2-7}
& $\mu^{\pm}_{k,i}$ & Peak amplitude & $2\times d_j$ (per sign) & \eqref{eq:wavelet_transform} & $[10, 50]$ & $50$ \\
& $\omega^{\pm}_{k,i}$ & Carrier frequency & $2\times d_j$ (per sign) & \eqref{eq:wavelet_transform} & $[0.3, 1]$ & $0.3$ \\
& $\ell^{\pm}_{k,i}$ & Spatial extent & $2\times d_j$ (per sign) & \eqref{eq:wavelet_transform} & $[10, 80]$ or $\eta_k/\omega^{\pm}_{k,i}$ & $80$ \\
& $\eta_k$ & Component-level extent scale & $1$ & \eqref{eq:wavelet_transform} & $[10, 24]$ & $10$ \\
\midrule
&\multicolumn{6}{l}{\textbf{Tensor Interference (Coupling)}} \\
\cmidrule{2-7}
& $\mu^{\pm}_{0,k,i}$ & Base cross-coordinate coupling strength & $2\times d_j$ (per sign) & \eqref{eq:tensor_interference} & $[10, 20]$ & $10$ \\
& $\mu^{\pm}_{k,i}$ & Effective amplitude (scaled as $\mu^{\pm}_{0,k,i}\,2^{d-1}$)\tnote{\textbf{$\S$}} & $2\times d_j$ (per sign) & --- & --- & Computed \\
& $\omega^{\pm}_{k,i}$ & Interference pattern spacing & $2\times d_j$ (per sign) & \eqref{eq:tensor_interference} & $[0.1, 0.7]$ & $0.7$\\
\midrule
&\multicolumn{6}{l}{\textbf{Radial Polynomial–Trigonometric (Coupling)}} \\
\cmidrule{2-7}
& $\mu_k$ & Isotropic modulation amplitude & $1$ & \eqref{eq:radial_hybrid_add} & $[0.4, 2]$ & $1$ \\
& $p_k$ & Radial envelope growth & $1$ & \eqref{eq:radial_hybrid_add} & $[0.4, 0.7]$ & $0.7$ \\
& $q_k$ & Ring spacing control & $1$ & \eqref{eq:radial_hybrid_add} & See note\tnote{\textbf{$\star$}} & $0.6$ \\
& $\omega_k$ & Ring density factor & $1$ & \eqref{eq:radial_hybrid_add} & See note\tnote{\textbf{$\star$}} & $10$ \\
\bottomrule
\end{tabular}

\begin{tablenotes}
\footnotesize
\item[$\dagger$] $d_j$ denotes the \emph{local} dimensionality of block $j$ (i.e., $d_j=|\mathcal I_j|$) to which component $k$ belongs. 
With a single block, $d_j=d$, the total dimension.
\item[$\ddagger$] Practical choice of the envelope rate $\gamma$ for Additive Periodic: on the default window $|a|\lesssim 100$, values $\gamma\in[2{\times}10^{-3},\,2{\times}10^{-1}]$ are typically effective. For $\gamma\gtrsim 0.2$, the envelope $E(t)=1-e^{-\gamma t}$ is $\approx 1$ over most of the domain (saturation); for $\gamma\ll 0.002$, the envelope ramps too slowly to reach full strength.
\item[$\S$] For Tensor Interference, the effective perturbation strength decays exponentially with dimension as $2^{-(d-1)}$ unless corrected. 
To maintain consistent behavior across dimensions, the user specifies a base amplitude $\mu^{\pm}_{0,k,i}$, which is automatically scaled as $\mu^{\pm}_{k,i}=\mu^{\pm}_{0,k,i}\,2^{d-1}$. 
This ensures that $\mu^{\pm}_{0,k,i}$ remains dimension-independent.
\item[$\star$] Radial polynomial–trigonometric presets: near-uniform rings use $q\in[0.9,\,1.2]$ with $\omega\in[0.1,\,0.4]$; widening rings use $q\in[0.4,\,0.6]$ with $\omega\in[5,\,10]$.
\end{tablenotes}

\end{threeparttable}
\end{table*}

\paragraph{Practical Exploration of Parameters} 
In addition, a supplementary document presents a curated set of plots that illustrate how individual parameters and their combinations shape the landscape geometry. 
Together, these resources bridge the gap between formal specification and intuitive understanding, supporting both reproducible research and hands-on exploration.

\paragraph{Practical Exploration of Parameters} 
While the mathematical formulations above define PORTAL's behavior precisely, users do not need to fully parse every equation to make effective use of the framework. 
In practice, what matters most is understanding the qualitative \emph{roles} and \emph{impacts} of parameters, that is, how they influence basin curvature, ruggedness patterns, and conditioning.
To facilitate this, PORTAL provides an interactive \emph{playground} environment where users can adjust parameters and immediately visualize their effects on the generated landscapes. 
The playground bridges the gap between formal specification and intuitive understanding, supporting both reproducible research and hands-on exploration. 
Comprehensive Python and MATLAB implementations of PORTAL, including the playground and all associated modules, are publicly available at  
\href{https://github.com/EvoMindLab/PORTAL}{\texttt{[https://github.com/EvoMindLab/PORTAL]}}.

\section{PORTAL Applications and Use Cases}
\label{sec:applications}

PORTAL's parametric design and layered architecture provide broad and fine-grained flexibility for constructing optimization landscapes.
Beyond traditional benchmark evaluation, PORTAL enables systematic generation of diverse problem instances with precise control over difficulty factors, supporting both established methodologies and emerging research directions.

We organize PORTAL's applications into four primary categories: 
\begin{itemize}
    \item \emph{Research Applications}: which include stress testing to identify algorithmic failure modes, controlled environments for algorithm development, and rigorous protocols for hypothesis-driven validation;
\item \emph{Specialized Suite Generation}, where PORTAL's configurability enables the design of benchmark suites that are fair, non-redundant, and diverse, with broad coverage of the instance space. Such suites can serve as standardized evaluation protocols for competitions, comparative studies, and domain-specific applications.
    \item \emph{Meta-Learning and Automation}: where PORTAL's capacity for large-scale, diverse dataset generation provides the foundation for training algorithm selection methods, automated algorithm design frameworks, and other data-driven approaches to optimization;
    \item \emph{Education and Training}: where PORTAL's interpretability and transparent parameter semantics make it a natural tool for demonstrating landscape characteristics, visualizing algorithm behavior, and supporting interactive learning environments.
\end{itemize}

Each category illustrates how PORTAL transforms benchmark generation into a versatile tool for both fundamental and applied research. The subsections below elaborate on these use cases and highlight how PORTAL supports systematic, reproducible, and scalable experimentation across the optimization research ecosystem.

\subsection{Research Applications: Systematic Algorithm Analysis}
\label{sec:stress_testing}

PORTAL's parametric design enables systematic algorithm analysis and stress testing, where individual problem characteristics can be isolated and progressively intensified to probe algorithmic robustness and failure modes. Unlike fixed benchmark suites, PORTAL allows researchers to vary one characteristic at a time while neutralizing others, ensuring that observed performance differences are attributable to the factor under investigation.

\subsubsection{Basin Curvature}
The exponent parameter $\mathfrak{p}_k$ controls the curvature of component basins, ranging from sublinear to superlinear growth. By neutralizing transformations, conditioning, and rotation, stress tests can isolate curvature effects and reveal how different algorithms respond to narrow valleys versus rapidly expanding basins.

\subsubsection{Conditioning}
Scaling parameters $\kappa_{k,i}^{\pm}$ directly determine the conditioning of each component. Systematic variation across logarithmic ranges produces instances from well-conditioned to severely ill-conditioned. Stress testing this axis highlights algorithm breakdown points as landscapes become progressively skewed.

\subsubsection{Variable Interaction}
The rotation matrices $\mathbf{R}_k$, parameterized by interaction specifications $\boldsymbol{\Psi}_k$, enable systematic control from fully separable to strongly coupled landscapes. Unlike earlier generators, PORTAL decouples separability from other features, allowing stress tests that specifically target sensitivity to interaction structures.

\subsubsection{Transformation-Induced Ruggedness}

PORTAL's transformation operators provide systematic stress testing of ruggedness by controlling both the \emph{amplitude} (depth/height of local optima) and the \emph{frequency} (spacing and density of optima). 
Different operators generate distinct topological patterns, such as ripples, concentric rings, or checkerboard interference, allowing evaluation of algorithmic robustness under diverse multimodality scenarios.
Importantly, while coupling transformations alter the variable interaction structure, they do not compromise the isolation of ruggedness effects: after transformation, each component can still be optimized dimension-wise. This enables controlled analysis of how algorithms handle ruggedness patterns of varying strength, density, and structural complexity.

\subsubsection{Asymmetry}
PORTAL also enables systematic asymmetry across all characteristics: curvature, conditioning, scaling, slopes, and local optima patterns can be independently configured for each dimension or direction. Asymmetric configurations introduce additional stress by breaking uniformity, forcing algorithms to adapt search behavior to heterogeneous landscapes and revealing weaknesses masked under symmetric settings.

\subsubsection{Multiple Components}
Stress testing with multiple components probes global search and deception resistance. PORTAL supports controlled construction of competitive basins, where high-quality local optima compete with the global optimum. Component placement and relative strength can be tuned to generate misleading attractors.

\subsubsection{Composition and Ill–Balancing}

PORTAL's support for block-wise composition facilitates several additional mechanisms for controlled experimentation:

\begin{itemize}
    \item \emph{Hybridization of challenges.} Different blocks can emphasize distinct characteristics, allowing researchers to study how algorithms balance progress across subfunctions when multiple challenges co-exist.
    \item \emph{Contribution imbalance.} Unequal weights $w_i$ allow some blocks to dominate the objective while others provide weaker or misleading signals. 
    This setting enables investigation of how algorithms allocate resources and adapt search pressure under imbalanced subfunctions.
\end{itemize}

\subsubsection{Scalability}
PORTAL supports systematic dimensional scaling, from moderate to very high dimensions. Block-wise composition further enables partially separable high-dimensional problems, maintaining interpretability while exposing scalability limits of algorithms under controlled separability structures.
This capability directly addresses the critical need for high-dimensional algorithm evaluation.

\subsubsection{Domain-Specific Benchmarking: Multimodal, Large-Scale, and Dynamic Optimization}

Beyond general optimization instances, PORTALs parameterization directly supports specialized subfields of optimization through appropriate configuration:

\begin{itemize}
    \item \emph{Multimodal optimization:} PORTAL's multi-component design enables the generation of landscapes with multiple global optima, providing benchmark instances for algorithms that must maintain diversity and locate all globally optimal solutions. Unlike fixed multimodal functions, the number, spacing, and relative quality of global optima can be explicitly controlled.
    \item \emph{Large-scale optimization:} Through partially separable structures and block-wise composition, PORTAL can generate high-dimensional problem instances with interpretable substructures. 
    \item \emph{Dynamic optimization:} By varying component parameters over time (e.g., positions, orientations, exponents, or transformation settings), PORTAL can mimic environmental changes with controllable intensity and frequency. This capability extends its use to the study of adaptive algorithms that must track moving optima in non-stationary environments.
\end{itemize}
These applications require no additional mechanisms beyond PORTAL's existing parameterization.

\subsection{Benchmark Suite Generation}
\label{sub:benchmark_generation}

A central capability of PORTAL is its role as an advanced and versatile \emph{generator} rather than a fixed catalog of problems. This distinction brings two key advantages.

First, PORTAL can generate a vastly diverse range of benchmark scenarios with precise, independent control over global characteristics such as basin curvature, conditioning, symmetry, variable interaction structures, transformation type and parameters, number and arrangement of components, block-wise composition (including weights), and dimensionality. 
These parameters are orthogonal and interpretable, enabling systematic exploration across the full spectrum of landscape features.

Second, for each scenario, PORTAL supports the creation of \emph{many non-identical instances} that share the same global characteristics but differ in low-level details. 
This is achieved by randomizing or permuting component centers, offsets, or angle assignments, while preserving the defining global settings. 
The result is sets of \emph{locally distinct but globally equivalent} instances, which are crucial for fair comparison across repeated runs, statistical testing, and the evaluation of deterministic algorithms without confounding effects from scenario drift.

Beyond these two core capabilities, PORTAL's flexibility enables coverage of the \emph{instance space} in a principled way. Methods such as instance space analysis and footprinting~\cite{smith2023instance} can be used to identify under-represented regions of the problem landscape (e.g., combinations of conditioning, interaction, and ruggedness that are missing in existing benchmarks). Because PORTAL can explicitly target these combinations, it can fill such gaps, yielding suites that are diverse, balanced, and methodologically rigorous.

These same advantages also make PORTAL directly applicable to specialized domains such as large-scale or multimodal optimization, where tailored suites can be generated by appropriate parameterization.  

Overall, PORTAL provides a systematic basis for \emph{benchmark suite generation}, ranging from general-purpose collections to domain-specific testbeds, supporting fair evaluation, algorithmic competitions, and targeted research campaigns designed around specific investigative goals.

\subsection{Meta-Algorithmic Applications}
\label{sub:meta_applications}

A growing frontier in optimization research lies in meta-algorithmics, including automated algorithm selection, hyperparameter tuning, and automated algorithm design.
These approaches require large, diverse datasets of optimization problems for training and validation, datasets that are both representative of the real problem space and systematically varied to prevent overfitting.
Traditional fixed-function suites are inherently inadequate for this purpose, as they provide only limited, static collections of instances.

PORTAL directly addresses this gap. Its parametric design enables:
\begin{itemize}
    \item \emph{Large dataset generation:} PORTAL supports the creation of extensive and highly diverse sets of problem instances through randomized exploration of its parameter space. 
    Parameters may also be stratified to enable structured random generation and balanced coverage of different challenge types.

    \item \emph{Unseen instance creation:} Because PORTAL's parameter space is continuous, it naturally supports hold-out strategies: training datasets can be drawn from one region, while test sets are sampled from disjoint regions, guaranteeing \emph{previously unseen} instances for testing generalization.
    \item \emph{Systematic diversity and coverage:} Features extracted via Exploratory Landscape Analysis (ELA) or Instance Space Analysis (ISA) can guide sampling strategies. Identified gaps in feature space coverage can be filled by generating targeted PORTAL instances, ensuring balanced and comprehensive datasets for machine learning.
    \item \emph{Controlled complexity progression:} Difficulty factors such as conditioning, ruggedness, or interaction strength can be gradually increased, enabling curriculum-style datasets for training adaptive meta-algorithmic methods.
\end{itemize}

\subsection{Educational Applications}
\label{sub:education}

Beyond research and meta-algorithmic applications, PORTAL provides significant value as an educational tool. Its clear parameter semantics and systematic control over landscape features make it well-suited for teaching optimization concepts in controlled, interactive environments.

Instructors can use PORTAL to construct tailored landscapes that highlight fundamental properties such as basin curvature, conditioning, separability, or multimodality, with each property introduced in isolation. The framework's interactive playground enables real-time parameter adjustment and visualization, allowing students to immediately observe how changes in exponents, scaling factors, or transformation parameters affect landscape structure and algorithmic behavior.

PORTAL supports comparative demonstrations by enabling instructors to generate identical problem instances for testing different algorithms, or to create systematic difficulty progressions for the same algorithm. Students can observe how different methods respond to controlled challenges, providing concrete insights into algorithmic trade-offs that are difficult to illustrate with fixed benchmark functions.

The interactive playground also facilitates hands-on learning exercises where students experiment with parameter effects. For example, adjusting ruggedness amplitude reveals its impact on both landscape topology and algorithm convergence, while manipulating rotation matrices illustrates how variable interactions alter search trajectories. This immediate visual feedback helps bridge the gap between theoretical understanding and practical algorithmic behavior, making abstract concepts tangible.

\section{Conclusion}
\label{sec:conclusion}

This paper introduced PORTAL, a comprehensive benchmark generation framework that addresses fundamental limitations of existing optimization test suites and generators. Through its systematic three-layer architecture and principled parameterization, PORTAL preserves interpretability and stability while enabling controlled variation in basin curvature, conditioning, separability, ruggedness, asymmetry, and compositional structure. 

By resolving entanglements in prior generators and supporting unbiased random sampling, PORTAL not only expands the diversity of attainable landscapes but also ensures reproducibility and comprehensive instance space coverage. 
These technical capabilities translate directly into practical impact: researchers can isolate or combine challenges for systematic algorithm analysis and failure mode identification, or design benchmark suites tailored to specific goals. 
They also enable the generation of rich datasets for meta-algorithmic studies and interactive educational demonstrations. 
In unifying flexibility, reproducibility, and breadth of coverage, PORTAL establishes a robust foundation for modern optimization research, particularly in emerging areas such as automated algorithm design and meta-learning.

\setstretch{0.96}
\small

\bibliography{bib}

@article{vermetten2025mabbob,
  title={MA-BBOB: A problem generator for black-box optimization using affine combinations and shifts},
  author={Vermetten, Diederick and Ye, Furong and B{\"a}ck, Thomas and Doerr, Carola},
  journal={ACM Transactions on Evolutionary Learning},
  volume={5},
  number={1},
  pages={1--19},
  year={2025},
  publisher={ACM New York, NY}
}

@article{bottou2018optimization,
  title={Optimization methods for large-scale machine learning},
  author={Bottou, L{\'e}on and Curtis, Frank E and Nocedal, Jorge},
  journal={SIAM review},
  volume={60},
  number={2},
  pages={223--311},
  year={2018},
  publisher={SIAM}
}

@article{li2023evaluation,
  title={Evaluation of frameworks that combine evolution and learning to design robots in complex morphological spaces},
  author={Li, Wei and Buchanan, Edgar and Le Goff, L{\'e}ni K and Hart, Emma and Hale, Matthew F and Wei, Bingsheng and De Carlo, Matteo and Angus, Mike and Woolley, Robert and Gan, Zhongxue and others},
  journal={IEEE Transactions on Evolutionary Computation},
  volume={28},
  number={6},
  pages={1561--1574},
  year={2023},
  publisher={IEEE}
}

@article{cao2022ai,
  title={Ai in finance: challenges, techniques, and opportunities},
  author={Cao, Longbing},
  journal={ACM Computing Surveys (CSUR)},
  volume={55},
  number={3},
  pages={1--38},
  year={2022},
  publisher={ACM New York, NY}
}

@article{Gandomi2023gnbg,
  title={A Generalized and Configurable Benchmark Generator for Continuous Unconstrained Numerical Optimization},
  author={Amir H. Gandomi and Mohammad Nabi Omidvar and Rohit Salgotra and Kalyanmoy Deb},
  journal={arXiv preprint arXiv:2312.07083},
  year={2023}
}

@article{munoz2014exploratory,
  title={Exploratory landscape analysis of continuous space optimization problems using information content},
  author={Mu{\~n}oz, Mario A and Kirley, Michael and Halgamuge, Saman K},
  journal={IEEE transactions on evolutionary computation},
  volume={19},
  number={1},
  pages={74--87},
  year={2014},
  publisher={IEEE}
}

@article{smith2023instance,
  title={Instance space analysis for algorithm testing: Methodology and software tools},
  author={Smith-Miles, Kate and Mu{\~n}oz, Mario Andr{\'e}s},
  journal={ACM Computing Surveys},
  volume={55},
  number={12},
  pages={1--31},
  year={2023},
  publisher={ACM New York, NY}
}

@article{bartz2020benchmarking,
  title={Benchmarking in optimization: Best practice and open issues},
  author={Bartz-Beielstein, Thomas and Doerr, Carola and Berg, Daan van den and Bossek, Jakob and Chandrasekaran, Sowmya and Eftimov, Tome and Fischbach, Andreas and Kerschke, Pascal and La Cava, William and Lopez-Ibanez, Manuel and others},
  journal={arXiv preprint arXiv:2007.03488},
  year={2020}
}

@article{jamil2013literature,
  title={A literature survey of benchmark functions for global optimisation problems},
  author={Jamil, Momin and Yang, Xin-She},
  journal={International Journal of Mathematical Modelling and Numerical Optimisation},
  volume={4},
  number={2},
  pages={150--194},
  year={2013},
  publisher={Inderscience Publishers Ltd}
}

@inproceedings{prager2022automated,
  title={Automated algorithm selection in single-objective continuous optimization: a comparative study of deep learning and landscape analysis methods},
  author={Prager, Raphael Patrick and Seiler, Moritz Vinzent and Trautmann, Heike and Kerschke, Pascal},
  booktitle={International Conference on Parallel Problem Solving from Nature},
  pages={3--17},
  year={2022},
  organization={Springer}
}

@inproceedings{whitley2002testing,
  title={Testing, evaluation and performance of optimization and learning systems},
  author={Whitley, D and Watson, Jean-Paul and Howe, Adele and Barbulescu, Laura},
  booktitle={Adaptive Computing in Design and Manufacture V},
  pages={27--39},
  year={2002},
  organization={Springer}
}

@inproceedings{shir2018compiling,
  title={Compiling a benchmarking test-suite for combinatorial black-box optimization: a position paper},
  author={Shir, Ofer M and Doerr, Carola and B{\"a}ck, Thomas},
  booktitle={Proceedings of the Genetic and Evolutionary Computation Conference Companion},
  pages={1753--1760},
  year={2018}
}

@Techreport{hansen2009real,
  title={Real-parameter black-box optimization benchmarking 2009: Noiseless functions definitions},
  author={Hansen, Nikolaus and Finck, Steffen and Ros, Raymond and Auger, Anne},
  year={2009},
  institution  ={INRIA}
}

@Techreport{chen2014problem,
  title={Problem definitions and evaluation criteria for {CEC} 2015 special session on bound constrained single-objective computationally expensive numerical optimization},
  author={Chen, Qin and Liu, B and Zhang, Qiang and Liang, Jing and Suganthan, Ponnuthurai and Qu, Boyang},
  institution  ={Technical Report, Computational Intelligence Laboratory, Zhengzhou University, Zhengzhou, China and Technical Report, Nanyang Technological University},
  year={2014}
}

@Techreport{awad2016problem,
  title={Problem definitions and evaluation criteria for the {CEC} 2017 special session and competition on single objective bound constrained real-parameter numerical optimization},
  author={Awad, N H and Ali, M Z and Liang, J J and Qu, B Y and Suganthan, P N},
  year={2016},
  institution  ={Nanyang Technological University Singapore}
}

@article{hansen2021coco,
  title={{COCO}: A platform for comparing continuous optimizers in a black-box setting},
  author={Hansen, Nikolaus and Auger, Anne and Ros, Raymond and Mersmann, Olaf and Tu{\v{s}}ar, Tea and Brockhoff, Dimo},
  journal={Optimization Methods and Software},
  volume={36},
  number={1},
  pages={114--144},
  year={2021},
  publisher={Taylor \& Francis}
}

@article{zhao2023automated,
  title={Automated design of metaheuristic algorithms: A survey},
  author={Zhao, Qi and Duan, Qiqi and Yan, Bai and Cheng, Shi and Shi, Yuhui},
  journal={arXiv preprint arXiv:2303.06532},
  year={2023}
}

@article{liu2024systematic,
  title={A systematic survey on large language models for algorithm design},
  author={Liu, Fei and Yao, Yiming and Guo, Ping and Yang, Zhiyuan and Zhao, Zhe and Lin, Xi and Tong, Xialiang and Yuan, Mingxuan and Lu, Zhichao and Wang, Zhenkun and others},
  journal={arXiv preprint arXiv:2410.14716},
  year={2024}
}

@article{tornede2023algorithm,
  title={Algorithm selection on a meta level},
  author={Tornede, Alexander and Gehring, Lukas and Tornede, Tanja and Wever, Marcel and H{\"u}llermeier, Eyke},
  journal={Machine Learning},
  volume={112},
  number={4},
  pages={1253--1286},
  year={2023},
  publisher={Springer}
}

@Techreport{mohamed2020problem,
  title={Problem definitions and evaluation criteria for the {CEC} 2021 special session and competition on single objective bound constrained numerical optimization},
  author={Ali Wagdy Mohamed and Anas A Hadi and Ali K Mohamed and Prachi Agrawal and Abhishek Kumar and PN Suganthan},
  year={2020},
  institution  ={Nanyang Technological University Singapore}
}

@Techreport{yue2019problem,
  title={Problem definitions and evaluation criteria for the {CEC} 2020 special session and competition on single objective bound constrained numerical optimization},
  author={Yue, CT and Price, Kenneth V and Suganthan, Ponnuthurai N and Liang, JJ and Ali, Mostafa Z and Qu, BY and Awad, Noor H and Biswas, Partha P},
  institution  ={Comput. Intell. Lab., Zhengzhou Univ., Zhengzhou, China, Tech. Rep},
  year={2019}
}

@article{cheng2018solving,
  title={Solving Incremental Optimization Problems via Cooperative Coevolution},
  author={Cheng, Ran and Omidvar, Mohammad Nabi and Gandomi, Amir H and Sendhoff, Bernhard and Menzel, Stefan and Yao, Xin},
  journal={IEEE Transactions on Evolutionary Computation},
  volume={23},
  number={5},
  pages={762--775},
  year={2018},
  publisher={IEEE}
}

@inproceedings{branke1999memory,
    author        = "Juergen Branke",
    title         = "Memory enhanced evolutionary algorithms for changing optimization problems",
    booktitle       = "IEEE Congress on Evolutionary Computation",
    publisher =    "IEEE", 
    volume        = "3", 
    year          = "1999",
    pages         = "1875--1882",
}

@incollection{branke2003designing,
    author      = "Juergen Branke  and Hartmut Schmeck ",
    title      = " Designing Evolutionary Algorithms for Dynamic Optimization Problems",
    editor      = " A. Ghosh and S. Tsutsui ",
    booktitle   = "Advances in Evolutionary Computing",
    publisher   = "Springer Natural Computing Series",
    address     = "",
    year        = 2003,
    volume        = "",
    pages       = "239--262",
    chapter     = "",
}

@article{hussain2017common,
  title={Common benchmark functions for metaheuristic evaluation: A review},
  author={Hussain, Kashif and Salleh, Mohd Najib Mohd and Cheng, Shi and Naseem, Rashid},
  journal={JOIV: International Journal on Informatics Visualization},
  volume={1},
  number={4-2},
  pages={218--223},
  year={2017}
}

@Techreport{suganthan2005problem,
    author =       "P. N. Suganthan and N. Hansen and J. J. Liang and K. Deb and Y.-P. Chen and A. Auger and S. Tiwari",
    year =         "2005",
    title =        "Problem definitions and evaluation criteria for the CEC 2005 special session on real-parameter
optimization",
    institution   =  "Nanyang Technological University",
    type =         "",
    number =       "",
    address =      "",
    month =        "",
    note =         "",
}

@article{li2018open,
  title={An open framework for constructing continuous optimization problems},
  author={Li, Changhe and Nguyen, Trung Thanh and Zeng, Sanyou and Yang, Ming and Wu, Min},
  journal={IEEE Transactions on Cybernetics},
  volume={49},
  number={6},
  pages={2316--2330},
  year={2018},
  publisher={IEEE}
}

@article{yazdani2020benchmarking,
  author={Yazdani, Danial and Omidvar, Mohammad Nabi and Cheng, Ran and Branke, Jürgen and Nguyen, Trung Thanh and Yao, Xin},
  journal={IEEE Transactions on Cybernetics}, 
  title={Benchmarking Continuous Dynamic Optimization: Survey and Generalized Test Suite}, 
  year={2022},
  volume={52},
  number={5},
  pages={3380-3393},
  publisher={IEEE},
}

@article{yazdani2021DOPsurveyPartA,
  title={A Survey of Evolutionary Continuous Dynamic Optimization Over Two Decades -- Part {A}},
  author={Yazdani, Danial and Cheng, Ran and Yazdani, Donya and Branke, J{\"u}rgen and Jin, Yaochu and Yao, Xin},
  journal={IEEE Transactions on Evolutionary Computation},
  volume={25},
  number={4},
  pages={609--629},
  year={2021},
  publisher={IEEE}
}

@article{yazdani2021DOPsurveyPartB,
  title={A Survey of Evolutionary Continuous Dynamic Optimization Over Two Decades -- Part {B}},
  author={Yazdani, Danial and Cheng, Ran and Yazdani, Donya and Branke, J{\"u}rgen and Jin, Yaochu and Yao, Xin},
  journal={IEEE Transactions on Evolutionary Computation},
  volume={25},
  number={4},
  pages={630--650},
  year={2021},
  publisher={IEEE}
}

@article{yazdani2021ieee,
  title={IEEE CEC 2022 competition on dynamic optimization problems generated by generalized moving peaks benchmark},
  author={Yazdani, Danial and Branke, Juergen and Omidvar, Mohammad Nabi and Li, Xiaodong and Li, Changhe and Mavrovouniotis, Michalis and Nguyen, Trung Thanh and Yang, Shengxiang and Yao, Xin},
  journal={arXiv: 2106.06174},
  year={2021}
}

@article{li2013benchmark,
  title={Benchmark functions for the CEC 2013 special session and competition on large-scale global optimization},
  author={Li, Xiaodong and Tang, Ke and Omidvar, Mohammad N and Yang, Zhenyu and Qin, Kai and China, Hefei},
  journal={gene},
  volume={7},
  number={33},
  pages={8},
  year={2013}
}

@article{li2013multimodalbenchmark,
  title={Benchmark functions for CEC’2013 special session and competition on niching methods for multimodal function optimization},
  author={Li, Xiaodong and Engelbrecht, Andries and Epitropakis, Michael G},
  journal={RMIT University, evolutionary computation and machine learning Group, Australia, Tech. Rep},
  year={2013}
}

@article{tang2007benchmark,
  title={Benchmark functions for the CEC’2010 special session and competition on large-scale global optimization},
  author={Tang, Ke and Li, Xiaodong and Suganthan, Ponnuthurai Nagaratnam and Yang, Zhenyu and Weise, Thomas},
  journal={Nature inspired computation and applications laboratory, USTC, China},
  volume={24},
  pages={1--18},
  year={2007}
}
\bibliographystyle{IEEEtran}

\end{document}